\documentclass[preprint,12pt]{elsarticle}

\usepackage{graphicx}
\usepackage{amssymb}
\usepackage{amsmath}
\usepackage{algorithm}
\usepackage{algorithmic}
\usepackage{url}
\usepackage{caption}
\usepackage{subcaption}

\DeclareMathOperator{\EX}{\mathbb{E}}
\usepackage{lineno}
\usepackage{placeins}
\usepackage{hyperref}
\usepackage{amsmath,amsfonts,amsthm,bm} 
\hypersetup{pdfauthor={Haoran Su}}

\journal{Transportation Research Part C}

\begin{document}

\begin{frontmatter}


\title{Dynamic Queue-Jump Lane for Emergency Vehicles under Partially Connected Settings: A Multi-Agent Deep Reinforcement Learning Approach}


\author[1]{Haoran Su}
\author[1]{Kejian Shi}
\author[1]{Joseph. Y.J. Chow}
\author[1]{Li Jin\corref{cor1}}
\ead{lijin@nyu.edu}
\cortext[cor1]{Corresponding author}
\address[1]{Tandon School of Engineering, New York University, Brooklyn, NY, 11201}
\begin{abstract}
Emergency vehicle (EMV) service is a key function of cities and is exceedingly challenging due to urban traffic congestion. A main reason behind EMV service delay is the lack of communication and cooperation between vehicles blocking EMVs. In this paper, we study the improvement of EMV service under V2X connectivity. We consider the establishment of dynamic queue jump lanes (DQJLs) based on real-time coordination of connected vehicles in the presence of non-connected human driven vehicles. We develop a novel Markov decision process formulation for the DQJL coordination strategies, which explicitly accounts for the uncertainty of drivers' yielding pattern to approaching EMVs. Based on pairs of neural networks representing actors and critics for agent vehicles, we develop a multi-agent actor-critic deep reinforcement learning algorithm which handles varying number of vehicles and random proportion of connected vehicles in the traffic. Approaching the optimal coordination strategies via indirect and direct reinforcement learning, we present two schemata to address multi-agent reinforcement learning on this connected vehicle application. Both approaches are validated, on a micro-simulation testbed SUMO, to establish a DQJL fast and safely. Validation results reveal that, with DQJL coordination strategies, it saves up to 30\% time for EMVs to pass a link-level intelligent urban roadway than the baseline scenario.
\end{abstract}
\end{frontmatter}


\section{Introduction}
Increasing population and urbanization have made it exceedingly challenging to operate urban emergency services efficiently. For example, historical data from New York City, USA \cite{NY2019} shows that the number of annual emergency vehicle (EMV) incidents has grown from 1,114,693 in 2004 to 1,352,766 in 2014, with corresponding average response time of 7:53 min and 9:23 min, respectively \cite{Emergency2014}. This means an approximately 20\% increase in response time in ten years. In the case of cardiac arrest, every minute until defibrillation reduces survival chances by 7\% to 10\%, and after 8 minutes there is little chance of survival \cite{Heart2013}. Cities are less resilient with worsening response time from EMVs (ambulances, fire trucks, police cars), mainly due to traffic congestion.

The performance of these EMV service systems in congested traffic can be improved with technology. 
As a core of modern ITSs, wireless vehicle-to-everything (V2X) connectivity, such as 5G-cellular network recently, provide significant opportunities for improving urban emergency response. On the one hand, wireless connectivity provides EMVs the traffic conditions on possible routes between the station (hospital, fire station, police station, etc.) and the call, which enables more efficient dispatch and routing. Through V2I communication with the EMVs on duty, traffic managers can broadcast the planned route of EMVs to non-EMVs that may be affected, and non-EMVs can cooperate to form alternative queue-jump lanes for approaching EMVs.

In this study, we introduce a novel concept of dynamic queue-jump lanes (DQJL). A queue jump lane (QJL) \cite{Zhou2005Performance,Cesme2015Queue} is a lane type used for bus operation so that buses can bypass long queues prior to intersections. In the context of EMVs, we define a DQJL to be a dynamic lane that is formed while an EMV is en-route to clear the downstream space. This is achieved by using V2X technology to inform downstream vehicles of the temporary change in the lane type so that vehicles would shift out of the lane. Doing so should significantly reduce the response time for EMVs on the urban roadway. By monitoring and managing the connected components of the roadway, we are able to produce real-time coordination strategies to clear a path for the approaching EMVs.

DQJL involves several uncertainties that impact the its implementation. One is how far downstream should the V2X set the lane type? Setting the DQJL too far downstream would negatively impact the background traffic. Setting it too short might not give vehicles enough time to comply. Secondly, a DQJL strategy can be implemented with an allocation recommendation to inform downstream vehicles of which open spaces to pull over to. Open spaces are finite and limited, and their usage depends on drivers' compliance. Poorly coordinated allocations can lead to low compliance due to difficult maneuvering requirements. The latter decision is called "DQJL with coordination" while  DQJL without coordination indicates no allocation recommendation given. Consider the stochastic driving behaviors and other uncertainties in the road environment, we formulate the problem into a partially observable Markov decision process (MDP). Aiming for real-time decision making when coordinating, we adopt the multi-agent actor-critic deep reinforcement learning to approach optimal coordination strategies. The methodology considers partial connected settings, indicating our frameworks are flexible with varying numbers of vehicles and connected vehicles' penetration rates.

The main contributions of the presented study are as follows: we introduce the concept of DQJL and solve the objectives of DQJL applications under stochastic road environment; we develop a multi-agent actor-critic deep reinforcement learning algorithm which copes with varying numbers of agent vehicles and penetration rates; addressing the real-time optimal coordination strategies via indirect and direct approach, we validate the DQJL coordinated system performs significantly better than the baseline scenario without any coordination.

The rest of the paper is organized in the following sections. In Sec. \ref{sec:literature_review}, we overview related studies on queue-jump lanes for EMVs and actor-critic reinforcement learning methods in similar ITS applications. In Sec. \ref{sec:modeling}, we elaborate the DQJL problem formulation as a Markov decision process. In Sec. \ref{sec:methodology}, we demonstrate our methodology, including the indirect and direct multi-agent reinforcement learning approaches. We elaborate our experimental design and synthesize a simulation dataset for the training in Sec. \ref{sec:experiment}. We validate the proposed methodology and provide insights on the results in Sec. \ref{sec:validation}. At last, conclusions are made in Sec. \ref{sec:conclusion}.

\begin{figure}
    \centering
    \includegraphics[width=\textwidth]{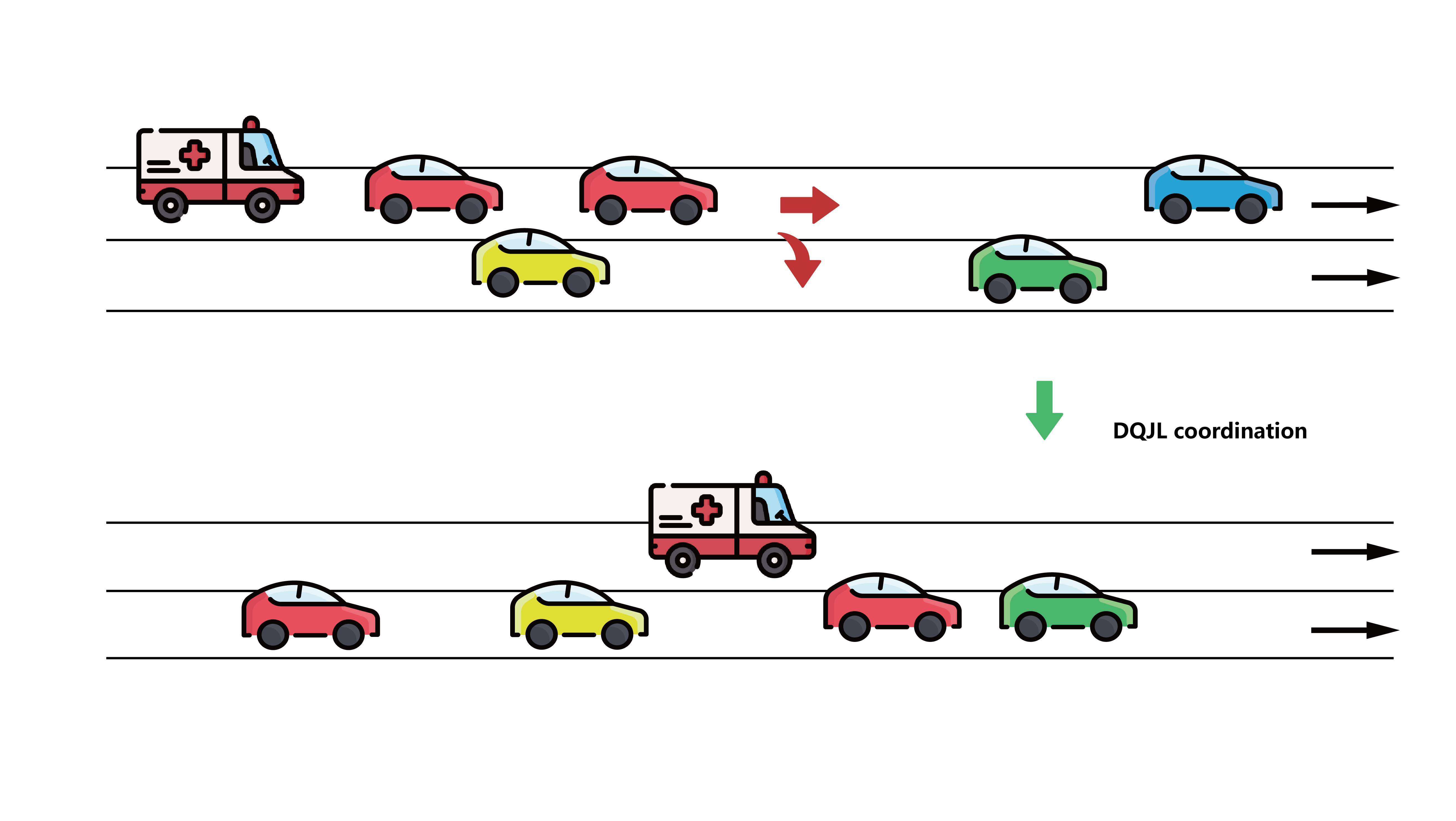}
    \caption{DQJL Coordination for an emergency vehicle.}
    \label{fig:dqjl_demo}
\end{figure}

\section{Literature Review}\label{sec:literature_review}
In this section, we discuss literature available on Queue jump lane in Subsec. \ref{sec:dqjl_literature}, and existing multi-agent deep reinforcement learning techniques on similar connected vehicle applications in Subsec. \ref{sec:drl_literature}.
\subsection{Queue jump lane for emergency vehicles}
\label{sec:dqjl_literature}
Queue jump lane (DQJ) has never been used for EMV deployment and is considered a novel operation strategy to apply this technology for EMV deployment with the aid of connected vehicle technologies.
Although QJL is a relatively new technology, literature has already demonstrates the positive effects they have in reducing travel time variability, especially when used in conjunction with transit signal priority (TSP). However, they are all based on moving-bottleneck models for buses \cite{Zhou2005Performance,Cesme2015Queue}; we are borrowing this bus operation strategy for EMV deployment in our setting, since EMVs typically move faster than non-EMVs and since EMVs can “preempt” non-EMV traffic because of their priority. 

At the same time, with only siren technologies, other vehicles often do not get enough warning time from the EMVs. Even then, there is a lack of clarity in the direction of the route to avoid. The confusion, particularly under highly congested scenarios, leads to increased delays as mentioned above, and also to 4 to 17 times higher accident rates \cite{Buchenscheit2009AVE} and increased severity of collisions \cite{Yasmin2012Effects}, which can further lead to increased response time. A study by Savolainen et al. \cite{Savolainen2010Effects} using data from Detroit, Michigan, verified this sensitivity of driver behaviors to different ITS communication methods for EMV route information. 

Hannoun et al. \cite{Hannoun2019Facilitating} use mixed integer linear programming to develop a static path clearance control strategy for approaching EMVs while QJLs have not been studied as a dynamic control strategy and there's an urgent need for capturing uncertainties in realistic traffic conditions, especially in events under non-deterministic setting such as yielding to an approaching EMV. To apply dynamic control strategy for vehicle's motion planning for QJL establishment, we introduce the concept of dynamic queue jump lane (DQJL), see Fig. \ref{fig:dqjl_demo}. During the process of clearing a QJL for the passing EMV, non-EMVs are constantly monitored and instructed with actions so that DQJLs are established quickly and safely. In the presented work, we also investigate the DQJL applications under a partially connected scenario, under which some vehicles are equipped with communication devices to establish complete communication channels between themselves and the infrastructure but others do not.

\subsection{Actor-critic method in ITS applications}\label{sec:drl_literature}
Many state-of-the-art ITS applications are adopting deep reinforcement learning techniques to realize connected vehicle applications. Among reinforcement learning schemes, the actor critic method, which combines the Q-learning and policy gradient, has been favored by extensive connected vehicle applications on cooperative traffic signal controls \cite{aslani2019developing, Chu2020, GUO2019313}, connected vehicle platooning \cite{wei2018design}, or trajectory planning \cite{yu2018intelligent, Yang2017Developments}.

Extending the actor critic algorithm from a single "super-agent" perspective to the multi-agent setting allows us to concentrate more on microscopic traffic management problems under connected settings. Under the multi-agent setting, it is assumed, by the essential definition of multi-agent systems that each vehicle doesn't have the full knowledge of the road environment, but can partially observe its surroundings. The local observation contains information obtained through physical surroundings, such as neighboring vehicles' positions, as well as messages through communication channels among connected vehicles and the infrastructure. Different actor critic architectures are used by these connected vehicle applications for various application purposes.

The fully centralized setting describes that all information is processed in a centralized controlled system. Namely, the policy and value networks for all vehicles, connected or autonomous, are stored in the centralized controller. All vehicles report their own observation spaces to the controller and, as a result, the controller has the full knowledge of the road environment. Decisions for each connected vehicles are then made by the centralized controller. The fully centralized setting based on actor critic methods appear in vehicle communication establishment \cite{Sahin2018Reinforcement}, coordination through intersections \cite{guan2020centralized}, and a previous study on DQJL coordination under fully connected scenario \cite{su2020v2i}. However, the fully centralized setting does not accommodate this application due to its limited usage under partially connected road environment, restricting the application's practicality within the short to medium future. The centralized setting also raises concerns on the high latency and synchronization cost, making real-time decision extremely difficult among vehicles.

At the same time, the fully decentralized settings are broadly approved in ITS applications. Each vehicle has its own policy network and actor network deployed on board. Each vehicle interacts with the environment independently. Methodologies addressing the multi-agent vehicle applications under this setting originate from the independent actor-critic (IAC) \cite{Tan93multi-agentreinforcement}. Plenty of robotic motion planning vehicle applications \cite{puccetti2019actor, huang2017parameterized} are based on the fully decentralized setting. The centralized controller, under this scheme, is no longer needed. Communications among vehicles are significantly reduced. While treating other vehicles as parts of the road environment faces huge risk of not converging as the environment becomes non-stationary. Vehicles also have difficulties in identifying their roles in cooperative tasks. Each individual vehicle is required to equipped advanced sensors, such as Lidar and cameras, to fully observe the road environment, significantly increasing the establishment cost of the system.

Proposed by \cite{lowe2017multiagent, foerster2017counterfactual}, the centralized training with decentralized execution framework overcomes challenges in previous settings. Under this framework, each vehicle has its own policy network deployed on board, while the corresponding value network is stored in the centralized controller. During the training stage, the centralized controller updates critic through TD learning with all agents' observation and their actions. The policy network then outputs the optimal action based on the vehicle's local observation. After the training stage, the value networks are completely abandoned and vehicles are able to operate based on their own observation spaces of the road environment and their policy networks, eliminating the concerns on synchronization. Since the centralized controller train value networks together, the road environment becomes stationary because the joint action is known, even though policy networks may change during the training stage.

The comparisons between these frameworks are summarized in Table. \ref{tab:marl_frameworks}. Denoting the local observation and action for a vehicle as $o^{i}_{t}$ and $a^{i}_{t}$ at step $t$, the joint observation is then denoted as $\mathbf{o}_{t}$ and joint action as $\mathbf{a}_{t}$. The actors and critics indicate the input information are either local or global. The latency column indicates whether the low latency and synchronization cost are highly required during execution stage or not, while the sensor column indicates if the vehicles need to equip the Lidar/cameras under the corresponding framework.
\begin{table}
\centering
\begin{tabular}{|c|c|c|c|c|}
\hline
Settings & Actor & Critic & Latency & Sensors \\ \hline
Centralized & $\pi(a^{i}_{t}|\mathbf{o}_{t}; \theta_{i})$ & $Q(\mathbf{o}_{t}, \mathbf{a}_{t};w_{i})$ & + & -\\ \hline
Decentralized & $\pi(a^{i}_{t}|o^{i}_{t}; \theta_{i})$  & $Q(o^{i}_{t}, a^{i}_{t};w_{i})$ & - & +\\ \hline
CTDE & $\pi(a^{i}_{t}|o^{i}_{t}; \theta_{i})$  &  $Q(\mathbf{o}_{t}, \mathbf{a}_{t};w_{i})$ & - & -\\ \hline
\end{tabular}
\caption{Summary of actor-critic based frameworks for connected vehicles.}
\label{tab:marl_frameworks}
\end{table}
\FloatBarrier

Multi-agent actor critic algorithms based on CTDE architecture, such as MADDPG, are applied in vehicle motion planning through signalized intersection application \cite{wu2020cooperative}, traffic flow optimization \cite{Cao2020Multi}, and vehicle network resources allocation \cite{kwon2019multi}. The existing literature view vehicles as the agents for cooperative or competitive traffic tasks. While these existing literature neither handle the heterogeneity of vehicle groups under partially connected setting, nor do they consider for varying size of the agents, i.e. vehicles. They also do not explicitly elaborate how to cope with discrete action space using multi-agent actor critic algorithm.

\section{Model Formulation} \label{sec:modeling}
In this section, we elaborate how we model the DQJL application as a partially-observed Markov decision process. In Subsec. \ref{subsec:dqjl_framework}, we reveal the structure of the DQJL coordination framework. The partially connected setting is explained in Subsec. \ref{subsec:partially_connected}, and components of the proposed MDP are illustrated in \ref{subsec:agents_states}, \ref{subsec:observation}, \ref{subsec:action}, \ref{subsec:reward} respectively. The objective function for the proposed MDP is presented in Subsec. \ref{subsec:objective_function}.

\subsection{DQJL coordination framework}\label{subsec:dqjl_framework}
In order to model the establishment of DQJL for an emergency vehicle (EMV), let us take a look at a typical urban road segment. An urban road segment consists of two lanes facing the same direction. When an EMV on duty approaches this road segment, the centralized control system based on cellular network will send out real time yielding coordination instructions to connected vehicles. 

With the cellular-assisted network prevailing in V2X communication \cite{SANTA20082850, MILANES201085, MUHAMMAD201850} nowadays, connected vehicles' information including positions, velocity, acceleration/deceleration and vehicle unique attributes such as vehicle length and most comfortable deceleration are communicated directly among connected vehicles and the infrastructure.

\begin{figure}[ht]
    \centering
    \includegraphics[width=\textwidth]{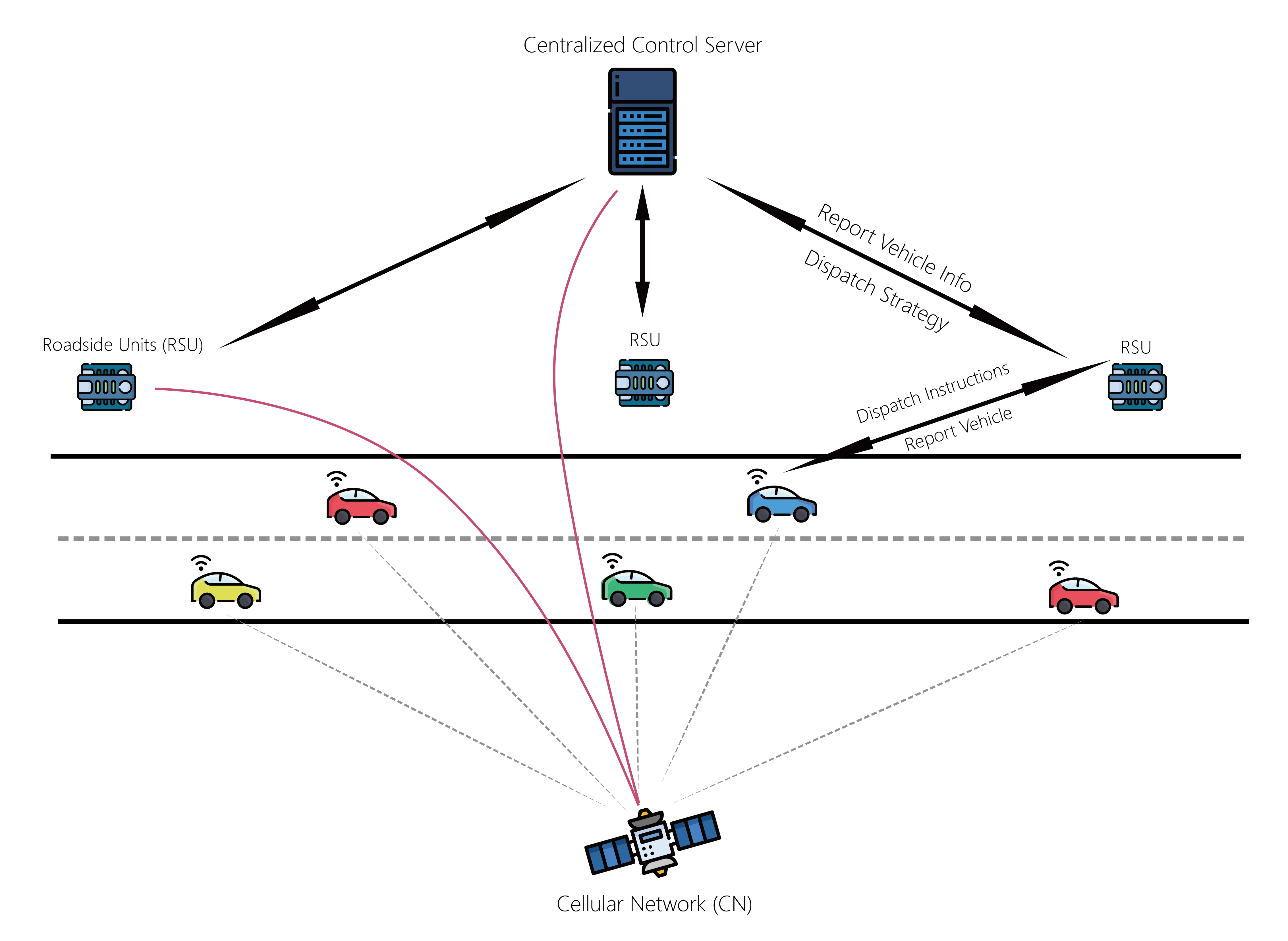}
    \caption{DQJL coordination framework under cellular-assisted vehicle networks. }
    \label{fig:DQJL_framework}
\end{figure}
\FloatBarrier

Approaching EMVs remain equipped with siren technologies to provide acoustic and visual warnings to vehicles, bikers and pedestrians. At the same time, EMVs' VANET-based [7] onboard warning systems are able to broadcast emergency vehicles' kinematic information to connected vehicles on the roadway. Sensors and detectors are deployed along the roadway to capture necessary features of non-connected vehicles. Vehicles' kinematic information including real time velocities and positions on the roadway are collected via technologies such as inductive and magnetic loop sensors \cite{scarzello2001vehicle, cheung2005traffic}.

Along the roadway, one or more roadside units (RSU) monitor and identify whether the approaching vehicles are connected or not. They also serve to synchronize data and coordinate when dealing with emergency cases. Information from the connected vehicles via the cellular-assisted vehicle network and data collected from sensors are processed together in RSUs. The RSUs then send the processed data to the centralized controller and wait for real-time coordination strategies as feedback. The RSUs then broadcast coordination instructions to the corresponding vehicles. The overall coordination framework is summarized in Fig. \ref{fig:DQJL_framework}. At the same time, the communication framework has a strict demand for low latency and synchronization cost in communication to ensure the compatibility of real-time coordination. 

To ensure the approaching EMV can pass the road segment as fast as possible, it is assumed the EMV does not perform lane change during passing so its average velocity can be maximized. The lane on which EMV is operating is named as the passing lane for referring convenience. All vehicles originally on the passing lane will pull over onto the other lane, which is referred as the neighboring lane.

\begin{figure}[!t]
    \centering
    \includegraphics[width=\textwidth]{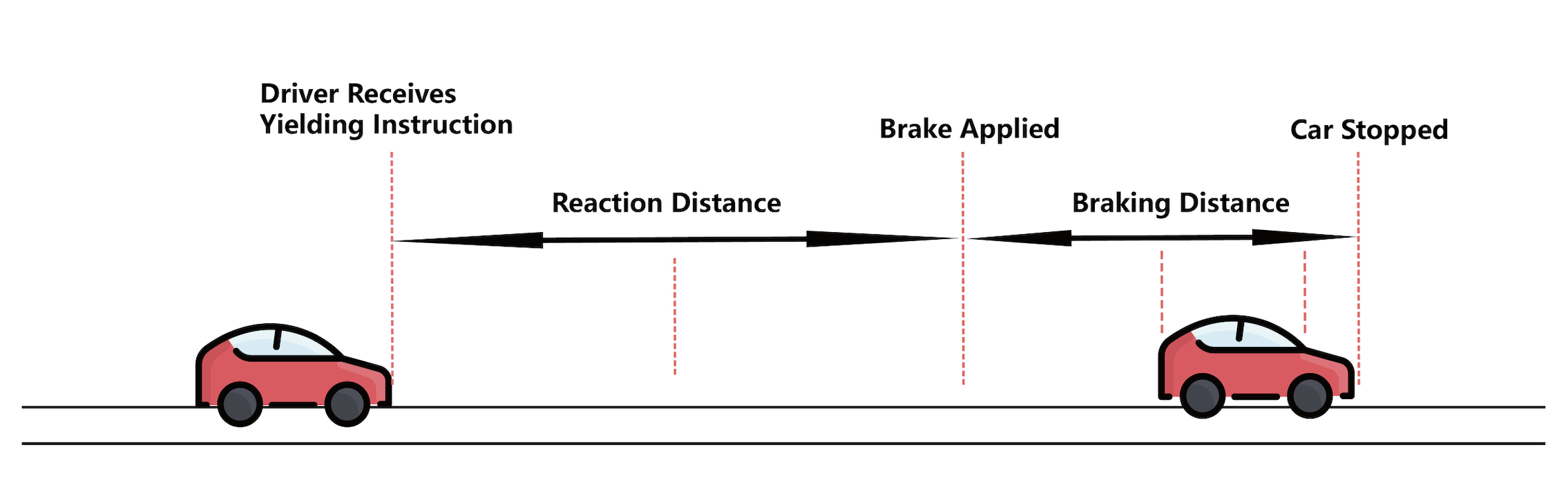}
    \caption{Braking process of a non-EMV after receiving a yielding instruction.}
    \label{fig:braking}
\end{figure}

Stochastic road dynamics originate from stochastic driving behaviors. For example, the uncertainties in drivers' perception-reaction time result in different reaction distance as shown in Fig. \ref{fig:braking}. Distinct braking abilities of different vehicles bring significantly different braking distances even when they start to decelerate at the same velocity. Stochastic driving behaviors also includes the randomness in lane-changing pattern when performing pull-over and $etc.$

Due to the uncertainties mentioned above, it is beneficial to picture the controlled system in a Markov decision process (MDP) framework so that the randomness can be addressed properly. Table. \ref{tab:MDP_notations} provides notations for variables used for describing the controlled system.
\begin{table}[!ht]
	\caption{Roadway Environment Notations}\label{tab:MDP_notations}
	\begin{center}
		\begin{tabular}{l l l l}
			Notation & Meaning \\\hline
			$x$  & front bump position of the vehicle  \\
			$v$  & velocity of the vehicle\\
			$l$ & length of the vehicle\\
			$b^{*}$ & baseline acceleration/deceleration of the vehicle\\
			$y$  & lane position of the vehicle \\
			$\xi$ & the yielding status of the vehicle\\
			$\phi$ & vehicle category: \{CV, HV or trivial vehicle\}\\
			$L$     & length of the road segment\\
			$d$     & minimum safety gap between two vehicle\\
			\hline
		\end{tabular}
	\end{center}
\end{table}

\subsection{The partially connected mixed traffic}\label{subsec:partially_connected}
According to the description of the controlled system above, the stream of vehicles has two categories of vehicles: the first one is the traditional human-driven vehicles (HV) with no advanced communication equipment and automation technologies; The second type is the connected vehicles (CV), which are highly connected within the framework and are able to share real-time information to the infrastructure or other CVs. For this particular application, vehicular automated technologies that operate the wheeling or acceleration of the vehicle are not considered, which means all vehicles are essentially operated by human drivers. Inspired by control strategies for partially automated traffic \cite{lazar2019learning, Biyik2019The}, the study develops a set of control strategies for the partially connected traffic.

On the one hand, HVs and CVs both contribute to the stochastic road dynamics due to the fact that they are operated by human drivers. All vehicles are unique with respect to the vehicles length and their deceleration rate as a result of their distinct braking abilities. All drivers are unique with respect to their perception reaction time, lane-changing behavior and other driving styles. These uncertainties constitute the stochastic traffic dynamics in this link-level segment.  

Moreover, CVs with explicit communication channels can be monitored and "managed" collectively to achieve cooperative tasks, even in the presence of HVs. The cooperation scheme is supposed to handle with the stochastic road dynamics mentioned above. At the same time, although HVs are not spontaneously sharing information with CVs or the intelligent infrastructure, their state information is constantly monitored by the infrastructure and observed by their surrounding CVs. 

\subsection{Agents and state}\label{subsec:agents_states}
In this MDP, the agents have two categories: HVs and CVs. The CVs are referred as active agents since their behaviors are controlled by the learning process, and HVs as passive agents since their behaviors are not controlled by the learning process.
At step $t$, the state representation of the an non-EMV, CV or HV, is expressed by the following features:
\begin{equation*}
     s^{i}_{t} = [x^{i}_{t}, y^{i}_{t}, v^{i}_{t}, \xi^{i}_{t}, l^{i}, b^{*}_{i}, \phi^{i}], 
\end{equation*}
where $\xi^{i}_{t}$ indicates whether or not a non-EMV is in the yielding process for the approaching EMV, and $\phi^{i}$ helps identify whether or not the $i$th non-EMV is a CV, HV or trivial vehicle.

Therefore, the ground truth state of the whole controlled system at step $t$ is characterized as 
\begin{equation*}
     \mathbf{s}_{t} = [s^{0}_{t}, s^{1}_{t}, s^{2}_{t}, \dots, s^{n}_{t}],
\end{equation*}
where $s^{0}_{t}$ is the state representation of the EMV and there are $n$ non-EMVs on the roadway when coordination process begins.

Notice that all non-EMVs and their drivers are unique and non-exchangeable. Non-EMVs are differentiated by their vehicle length $l$ and their baseline acceleration/deceleration $b^{*}$ due to their unique braking ability. The drivers are also differentiable due to nature of everyone's unique driving style and behavior. Thus, the agents are heterogeneous in this MDP framework.

\subsection{Observation space}\label{subsec:observation}
\begin{figure}
    \centering
    \includegraphics[width=\textwidth]{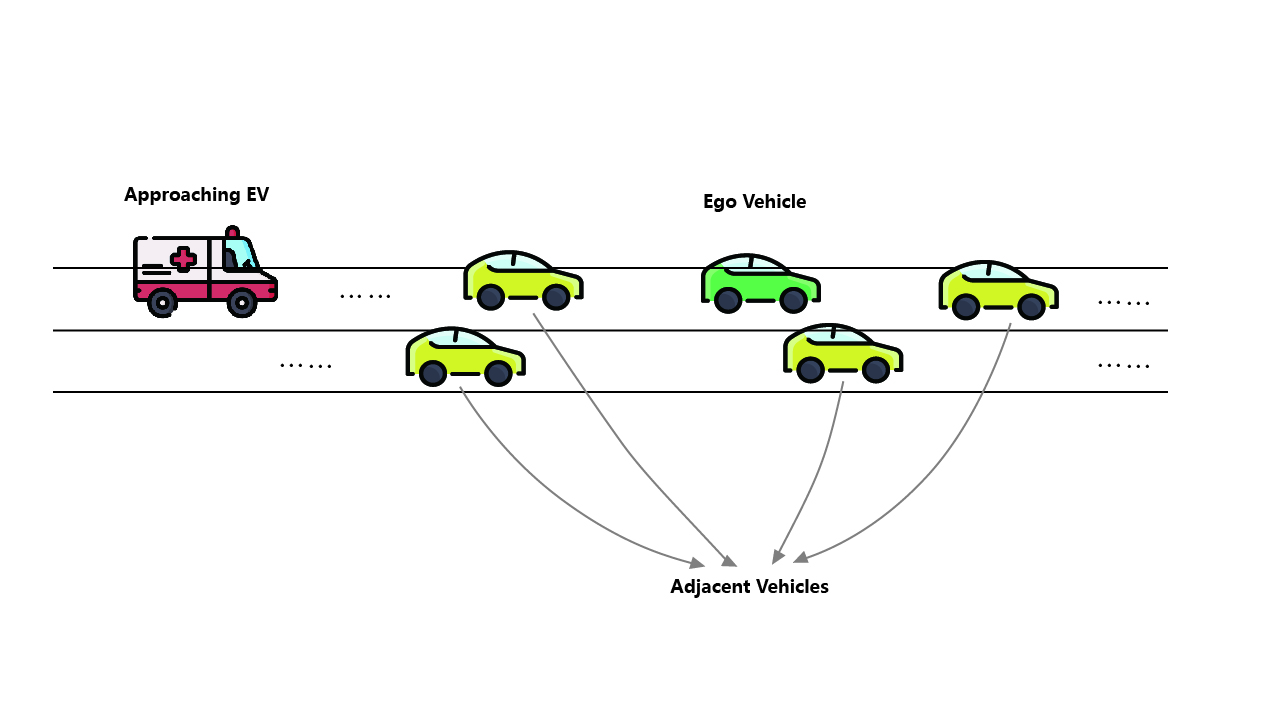}
    \caption{Example of the local observation captured by any non-EMV on the road segment.}
    \label{fig:local_observation}
\end{figure}

Each non-EMV only observes partial road segment. To define local observation of anon-EMV, the four closest adjacent non-EMVs are selected, which are the ego vehicle's leading vehicle, its following vehicle and the leading vehicle and the following vehicles on the neighboring lane. At the same time, the approaching EMV can broadcast its real time kinematic information to all CVs on this road segment, and any non-EMV would include the EMV's information in its own local observation. The adjacent vehicles and the approaching EMV imposes significant impacts on the ego vehicle's motion planning decisions. Fig. \ref{fig:local_observation} demonstrates the local observation captured by any non-EMV on this road segment.

For the $i$th non-EMV on this road segment, its local observation is described as
\begin{equation*}
    o^{i}_{t} = [s^{0}_{t}, s^{i}_{t}, s^{ij}_{t}, s^{ik}_{t}, s^{il}_{t}, s^{im}_{t}],
\end{equation*}
where $ij, ik, il, im$ represents $i$th vehicle's adjacent vehicles at step $t$ respectively, and $s^{0}_{t}$ stands for the state representation of the EMV at step $t$.

The dimension of any non-EMV's observation space should be fixed. If a non-EMV does not have a complete observation space, such as the leading vehicle of a lane, trivial vehicles are introduced to fill up the observation space. A trivial vehicle contains insignificant vehicle features and it is labeled during the learning process, which helps the controller to ignore its interaction with other vehicles.

\subsection{Action}\label{subsec:action}
For HVs, the actions are primarily determined by the distance between the EMV and itself when drivers hear the sirens or have visual confirmation of the vehicle. This distance-based action function with w.r.t to step is described as
\begin{equation}
    \label{eq:HV_reaction}
    a^{i}_{t} = \begin{cases}
    0, & \text{if } x^{i}_{t} - x^{i}_{t} \geq L_{HV}.\\
    1, & \text{otherwise}.\\
\end{cases}
\end{equation}

For CVs, the agents are human drivers operating the connected vehicles, so the action instructed to each human driver is to yield or not. Thus, the discrete action space for agent $i$ at step $t$ is represented as $a^{i}_{t} \in \{0, 1\}$. 

Notice that both the yielding and the moving forward instructions are high-level motion planning representations, and they don't determine the vehicles' absolute velocities as well as accelerations. With yielding or moving forward instructions, human drivers drive according to the road environment dynamics in addition to their unique and stochastic driving styles. Defining the action in such way circumvents the difficulty of explicitly indicating the specific kinematics attributes as well as the stochastic driving behaviors associated. The joint action of $n$ non-EMVs at step $t$ is then represented as
\begin{equation*}
    \mathbf{a}_{t} = [a^{1}_{t}, a^{2}_{t}, \dots, a^{n}_{t}].
\end{equation*}

\subsection{Reward}\label{subsec:reward}
For the DQJL application, three reward functions are named to approach the purpose of establishing a DQJL safely as fast as possible. First of all, to avoid vehicle collision, a significant negative reward $P_{collision}$ for overlapping of any two neighboring vehicles' positions is imposed:
\begin{equation*}
    r^{collision}_{t} = \begin{cases}
    P_{collision}, & \text{ if }x^{i}_{t} + d < x^{j}_{t} - l^{j}, \forall i, j \text{ when } y^{i}_{t} = y^{j}_{t}.\\
    0, & \text{otherwise.}\\
    \end{cases}
\end{equation*}
To motivate non-EMVs to clear the path, the second reward function penalizes every step elapsed,
\begin{equation*}
    r^{elapsed}_{t} = \begin{cases}
    -1, & \text{ if }x^{0}_{t} - l^{0} < L.\\
    0, & \text{otherwise,}\\ 
    \end{cases}
\end{equation*}
where $l^{0}$ represents the vehicle length of the approach EMV.
The third reward function emphasizes the urgency of yielding for each non-EMV in the way of the EMV. Namely, 
\begin{equation*}
    r^{i}_{t} = \begin{cases}
    -\frac{P_{priority}}{x^{i}_{t} - x^{0}_{t}}, & \text{if } x^{i}_{t} \leq L \text{ and } y^{i}_{t} = 0,\\
    0, & \text{otherwise,}\\
    \end{cases}
\end{equation*}
where $P$ is a penalty constant. This definition prioritizes the yielding of non-EMVs which are closer to the EMV that are downstream from it. To summarize, the immediate step reward at step $t$ is 
\begin{equation*}
    R(t) = r^{collision}_{t} + r^{elapsed}_{t} + \sum_{i=1}^{n}r^{i}_{t}(1 - y_{i}^{t}).
\end{equation*}

Since the application is cooperative, the reward received by each agent is identical to the team reward, which is
\begin{equation*}
    R^{1}_{t} = R^{2}_{t} = \dots = R^{n}_{t} = R_{t}.
\end{equation*}

\subsection{MDP objective function}\label{subsec:objective_function}
Defining the long-term return for the team as $U_{t}$ at step $t$, the objective function for the DQJL coordination can be represented as
\begin{equation}\label{eq:objective_function}
    \max U_{0} = \sum^{\infty}_{k=0} \gamma^{k} R_{k}.
\end{equation}

Solving \eqref{eq:objective_function} not only returns the optimal coordination strategy but also reveals the minimum amount of time for DQJL establishment. Notice that since the immediate step rewards are affected by the stochastic road environment, the expected return reflects the uncertainties during the DQJL coordination process.
\section{Multi-Agent Actor-critic Method for Coordination Strategies}\label{sec:methodology}
In this section, we introduce the our approach for DQJL coordination strategies. Starting with Subsec. \ref{subsec:bellman_optimality} and \ref{subsec:single_agent}, we revisit the preliminaries of RL and the actor critic method. In Subsec. \ref{subsec:main_algo}, we show the step-by-step derivation of the multi-agent actor-critic method for DQJL coordination.
In Subsec. \ref{subsec:indirect_approach}, we elaborate how to construct a road dynamic model reflecting the realistic traffic condition, and then employ the multi-agent actor critic algorithms under centralized training with decentralized execution framework as the indirect approach. In Subsec. \ref{subsec:direct_approach}, we demonstrate the pipeline for the embedding of the MARL algorithm on Simulation of Urban Mobility (SUMO) as the direct approach. The deep neural network structures supporting the presented algorithm as function approximators are presented in Subsec. \ref{subsec:DNN_structure}.
\subsection{Bellman optimality}\label{subsec:bellman_optimality}
In the fully observable MDP, an agent has the full knowledge of the environment $s \in S$ at step $t$, and select an action $a$ according to its policy $\pi(a|s)$. The state transition occurs as $s_{t} \sim s_{t+1}$ and an immediate step reward is received in the form of $R_t = R(s_t, a_t, s_{t+1})$.

If the MDP has infinite-horizon, the sampled expected return under some policy is calculated as $R^{\pi}_{t} = \sum_{\tau = t}^{\infty}\gamma^{\tau - t}r_{t}$, where $\tau \in [0,1)$ is the discount factor. From there, Q-function under some policy is defined as
\begin{equation}
    \label{eq:Q_function}
    Q^{\pi}(s, a) = \EX{}[R^{\pi}_{t}|s_{t} = s, a_{t} = a].
\end{equation}
Hence, the optimal Q-function is obtained from $Q^{*} = \max_{\pi}Q^{\pi}$, yielding the optimal policy $\pi^{*}$. Under $\pi^{*}$, $a \in \arg\max_{a_{t+1}}Q^{*}(s, a_{t+1})$.

According to the Bellman's Equation \cite{bellman1954}, the optimal Q-function is solved by
\begin{equation}
\label{eq:bellman}
    Q^{*}(s, a) = R_{t} +\gamma\sum_{s_{t+1} \in S}p(s_{t+1}|s_{t}, a_{t})\max_{a_{t+1}\in A}Q(s_{t}, a_{t+1}).
\end{equation}

Since $R_t$ and $p(s_{t+1}|s_{t}, a_{t})$ are unknown to the agent, the agent has to approach the optimality by purely data-driven dynamic programming based on the collected experience $(s_t, a_t, R_t, s_{t+1})$.

\subsection{Single-agent actor-critic method}\label{subsec:single_agent}
Actor-critic method, introduced by \cite{konda2000actor}, combines the policy gradient algorithm with value-based learning method. State-of-the-art actor-critic based methods, such as DDPG \cite{lillicrap2019continuous} and soft actor-critic method \cite{haarnoja2019soft}, exhibit outstanding performance in a variety of control problems in stochastic processes, especially in the motion planning control of connected and autonomous vehicles. Consider an intelligent vehicle as the agent with a discrete action space, e.g. driving forward and backing up. The vehicle's state-value function under some policy $\pi$ is calculated as
\begin{equation*}
    V_{\pi}(s) = \sum_{a\in A} \pi(a|s)Q_{\pi}(a, s),
\end{equation*}
where $\pi(a|s)$ represents the policy function and $Q_{\pi}(a, s)$ refers to the state-action function. $V_{\pi}(s)$ estimates the vehicle's standing in the application. With the approximation power of neural networks, the policy function is parameterized with a set of trainable parameters as $\pi(a|s; \bm{\theta})$, and the state-action function is parameterized as $Q_{\pi}(a, s; \mathbf{w})$.

The state-value function is then parameterized as $V_{\pi}(s; 
\bm{\theta}, \mathbf{w})$. By sampling an action through the vehicle's current policy $a \sim \pi(\cdot| s; \bm{\theta})$, the corresponding next state $s_{t+1}$ and immediate step reward $R_{t}$ are observed. 

The value network, i.e the critic, is trained via the temporal difference (TD) learning. First, $Q(s, a; \mathbf{w})$ and $Q(s_{t+1}, a_{t+1}; \mathbf{w})$ are computed and a TD target is set as
\begin{equation}\label{eq:td_target}
    y_{t} = R_{t} + \gamma Q(s_{t+1}, a_{t+1}; \mathbf{w}).
\end{equation}
The purpose of setting up \eqref{eq:td_target} is to minimize the difference between the $y_{t}$ and $Q(s_{t}, a_{t}; \mathbf{w})$. Mathematically,
\begin{equation*}
    \min L(\mathbf{w}) = \frac{1}{2}[y_{t} -Q(s_t, a_t; \mathbf{w})]^2,
\end{equation*}
and $\mathbf{w}$ is updated by the gradient ascent method as
\begin{equation}\label{eq:gradient_ascent_w}
    \mathbf{w}_{t+1} = \mathbf{w}_{t} + \alpha \frac{\partial L(\mathbf{w})}{\partial \mathbf{w}}|_{\mathbf{w} = \mathbf{w_{t}}},
\end{equation}
in which $\alpha$ is the learning rate for the value network.

The policy network, i.e. the actor, is updated through the traditional policy gradient algorithm. Denoting the policy gradient of an action $a_{t}$ as $\mathbf{g}(a_{t}, \bm{\theta}) = \frac{\partial \log \pi(a|s; \bm{\theta}))}{\partial \bm{\theta}}Q(s_{t}, a;\mathbf{w})$, it is easy to derive the policy gradient with respect to the state-value function as
\begin{equation}
    \frac{\partial V(s; \bm{\theta}, \mathbf{w}_{t})}{\partial \bm{\theta}} = \EX_{A}[\mathbf{g}(A,\bm{\theta})].
\end{equation}
By randomly sampling an action through Monte Carlo simulation $a_{t} \sim \pi(\cdot| s_{t}; \bm{\theta}_{t})$, $\mathbf{g}(a_{t})$ is guaranteed unbiased. $\bm{\theta}$ is then updated through the stochastic gradient descent as
\begin{equation}\label{eq:stochastic_gd_theta}
    \bm{\theta}_{t+1} = \bm{\theta}_{t} + \beta \mathbf{g}(a, \bm{\theta}_{t}),
\end{equation}
where $\beta$ is the learning rate of the policy network. By setting up the applicable reward functions, a vehicle's critic will improve its rating ability with regard to the ground truth estimation. The actor, on the other hand, will make the optimal decision by the critic's standard, guiding the vehicle achieve the application purpose.

\subsection{Multi-agent actor-critic method in discrete action space}\label{subsec:main_algo}
\begin{figure}
    \centering
    \includegraphics[width=0.9\textwidth]{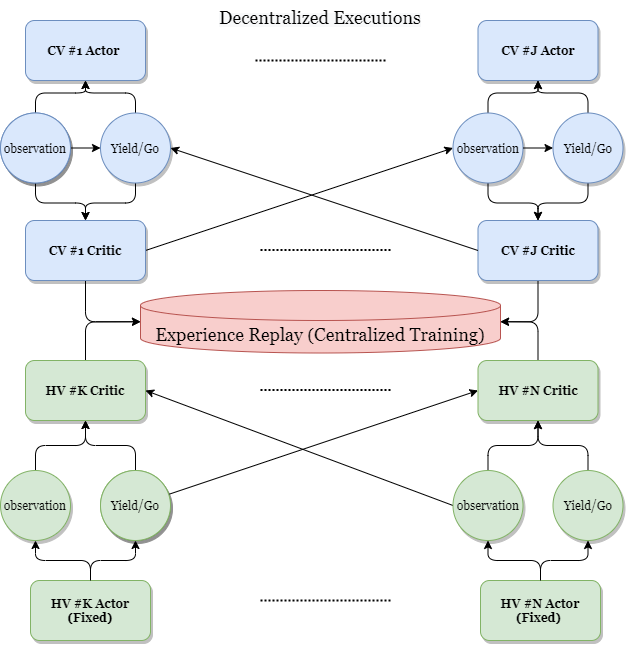}
    \caption{The CTDE architecture on the DQJL application.}
    \label{fig:DNN_architecture}
\end{figure}

Under the CTDE framework, each vehicle in the system, HV or CV, has a local policy network, and the corresponding is stored in the centralized controller. The focus of the learning process is on the policy networks of the CVs since HVs are not controlled by the learning process and their actors are fixed. To maintain the training consistency, HVs are viewed as agents. The main idea behind CTDE is that the road environment is considered stationary after knowing the joint actions of all non-EMVs even when some actors are updated in the training. Assuming there are $J$ CVs and total $N$ non-EMVs in the controlled system, the CTDE architecture is summarized in Fig. \ref{fig:DNN_architecture}.

The controlled system is described as a fixed length $L$ road environment with $n$ non-EMVs. These non-EMVs have parameterized policies with $\bm{\theta} = \{\theta_{0}, \theta_{1}, \dots, \theta_{n}\}$ and their policies are summarized as $\bm{\pi} = \{\pi_{1}, \pi_{1}, \dots, \pi_{n}\}$. $\theta$ for HVs are constant and they are not updated through training. Another set of parameters $\mathbf{w} = \{w_{1}, w_{2}, \dots, w_{n}\}$ is selected to denote the value networks stored in the centralized controller.

\subsubsection{Experience replay initialization}
Suppose at some step $t$, the joint observation of all vehicles is $\mathbf{s}_{t}$. A joint action $a_{t}$ is taken and the joint next state $\mathbf{s}_{t+1}$ is observed. An experience replay bank $D = [\mathbf{s}_{t}, \mathbf{s}_{t+1}, \mathbf{a}_{t}, R^{1}_{t}, R^{2}_{t}, \dots, R^{n}_{t}]$ to store the joint state, joint next state, joint action and joint reward so that centralized the training stage can have a more efficient use of the previous experience.
\subsubsection{Centralized critics update}
Similar to the employment of the TD learning shown in \eqref{eq:td_target} for single-agent RL, a TD target for the multi-agent scenario is named as
\begin{equation}
    \label{eq:td_target_multi}
    y_{t} = R_{t} + \gamma Q^{\bm{\theta}'}_{i}(\mathbf{s}_{t+1}, \mathbf{a}_{t+1};\mathbf{w}_{i}) |_{ a^{j}_{t+1} =  \theta_{j}'(o^{j}_{t+1})},
\end{equation}
and the target is to minimize the error between $y_{t}$ and $Q^{\bm{\theta}}_{i}(\mathbf{s}_{t}, \mathbf{a}_{t})$ as
\begin{equation}
    \label{eq:critic_update}
    \min L(\mathbf{w}_{i}) = \frac{1}{2}\EX_{(\mathbf{s}_t, a_t, R_t, \mathbf{s}_{t+1}) \in D}[(Q^{\bm{\theta}}_{i}(\mathbf{s}_{t}, \mathbf{a}_{t}) - y_{t})^{2}].
\end{equation}
In \eqref{eq:td_target_multi}, $\bm{\theta}' = \{\theta_1', \theta_2', \dots, \theta_n'\}$ represents the target policy set with delayed parameters. Through the gradient ascent algorithm as in \eqref{eq:gradient_ascent_w}, the value network considering for all agents' states and actions are updated.
\subsubsection{Decentralized actors update}
Actor deployed on each agent vehicle is updated through policy gradient. The gradient of expected return for $i$th non-EMV in the discrete action space is described as 
\begin{equation}
    \label{eq:policy_gradient}
    \nabla _{\theta_{i}}J(\theta_{i}) = \EX_{\mathbf{s}, a_i \sim \pi_i}[\nabla_{\theta_{i}}\pi_{i}(a^{i}_{t}|o^{i}_{t}) \nabla_{a^{i}_{t}}Q^{\bm{\pi}}_{i} (\mathbf{s}_{t}, \mathbf{a}_{t})],
\end{equation}
where $Q^{\bm{\pi}}_{i}(\mathbf{s}_{t}, \mathbf{a}_{t})$ refers to the action-value function stored in the centralized critic shown in \eqref{eq:critic_update}. The centralized critic value considers all actions from all agents, making the environment stationary in training. Since the action space is discrete, a Gumbel-softmax function \cite{gumbel_softmax} is employed to produce differentiable action samples from the policy when performing the Monte Carlo simulation. The differentiable sampled actions help to train the policy without using the log derivative trick as used in DDPG \cite{lowe2017multiagent}.

Notice in \eqref{eq:critic_update} that even though the experience replay doesn't explicitly memorize the joint next action, the centralized critic utilizes all agents' target actors to predict the joint next action and improve the stability of the learning during the update. The training of the centralized critics enhance the core idea of CTDE that, knowing the joint action, the road environment is stationary even though some policies change.

Following the idea of multi-agent actor critic method in discrete action space, we expand and summarize the proposed MARL algorithm in Algo. \ref{algo:training} for the optimal DQJL coordination strategy with dynamic amount of non-EMVs in the presence of HVs.
\begin{algorithm}[!ht]
\caption{Multi-Agent Actor Critic for DQJL Coordination}
\begin{algorithmic}[1]\label{algo:training}
\STATE Initialize $M$ pairs of policy networks and value networks
\STATE Initialize the experience history $D$ with mini-batch size
\FOR{each coordination training episode}
    \STATE reset road environment for the initial state 
    $\mathbf{s}$
    \STATE insert trivial vehicles until there are $M$ non-EMVs
    \FOR{each coordination training step}
        \STATE sample an action $a^{i}_{t}$ for each CV as $a^{i}_{t} = \pi(\cdot|o^{i}_{t})$
        \STATE sample an action $a^{i}_{t}$ for each HV based on \eqref{eq:HV_reaction}
        \STATE execute $\mathbf{a}_{t}$ and observe $\mathbf{s}_{t+1}$ and $R_{t}$
        \STATE store $(\mathbf{s}_{t}, \mathbf{a}_{t}, R_{t}, \mathbf{s}_{t+1})$ into $D$
        \STATE update state $\mathbf{s}_{t} \xleftarrow{} \mathbf{s}_{t+1}$
        \FOR{non-EMV = $1$ to $M$}
        \IF{the vehicle is non-trivial}
            \STATE draw a mini-batch sample $(\mathbf{s}_{t}, \mathbf{a}_{t}, R_{t}, \mathbf{s}_{t+1})$ from $D$ 
            \STATE set $y_{i}$ and update the critic according to \eqref{eq:critic_update}
            \IF{the vehicle is a CV}
            \STATE update the actor according to \eqref{eq:policy_gradient}
            \ENDIF
        \ENDIF
        \ENDFOR
        \STATE update target networks by $\theta_{i}' \xleftarrow{} (1-\tau)\theta_{i}' + \tau \theta_{i}$
    \ENDFOR
\ENDFOR
\end{algorithmic}
\end{algorithm}
\subsection{Deep neural network architecture}\label{subsec:DNN_structure}
Vehicle trajectories are temporal-sensitive data, and vehicles maintain action consistency if they are yielding to an approaching EMV, i.e. a yielding vehicle decelerates and changes lane until it parks into the neighboring lane. Utilizing a long short-term memory (LSTM) layer as the output layer for the actor, the network can further reduce non-stationarity. The action space for this application is binary, so the LSTM layer is followed by a Gumbel softmax function to output the probabilities of yielding or driving forward respectively.

As introduced in Algo. \ref{algo:training}, the proposed methodology can handle various number of vehicles in the system by fixing the dimension of state representations. Adopting a similar technique in a previous study \cite{su2020v2i}, trivial vehicles are added into the controlled system until there are $M$ pairs of networks. Although some networks are idle through the process, this technique circumvents the challenge in many connected vehicles applications where the number of agents is varying. This technique ensures the proposed methodology's engineering practicality.

\begin{figure}                                                 
    \centering
    \includegraphics[width=\textwidth]{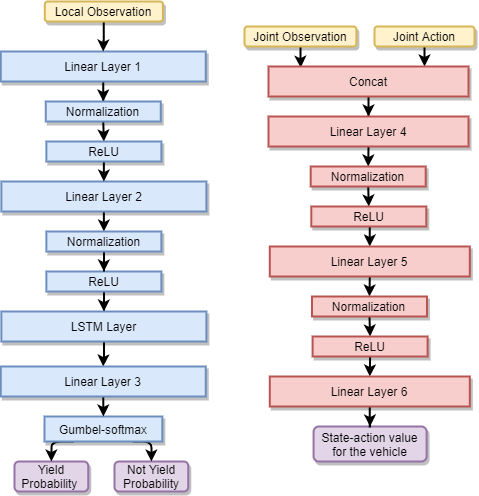}
    \caption{Deep neural network architectures for a pair of policy network and value network.}
    \label{fig:DNN}
\end{figure}

See Fig. \ref{fig:DNN} for the detail neural network architectures and dimensions of the policy network and value network. The policy network takes input of the vehicle's local observation, and processed through a double layer multi-layer perceptron (MLP). Each layer of the MLP consists of a normalization layer, a linear layer as well as a ReLU function. A LSTM layer is then attached, and two probabilities are resulted from the Gumbel-softmax function, standing for the vehicle's probabilities of yielding or not at this step respectively.

The value network takes the input of the road environment state representation and the joint action at this step. After concatenating them together, the network feeds the processed vector through a three layer MLP. The result from the value network is a scalar, representing the centralized state-action value for this particular agent. Adam Optimizer is selected to optimize losses for both networks.

For the policy network, the input is the flatten local observation with dimension of $6\times7$, linear layer 1 has dimension of $42\times64$ and linear layer 2 has dimension of $64\times128$. The LSTM layer takes input dimension of 128 and output 128. Linear layer 3 has dimension of $128 \times 2$. For the value network, the concatenation layer has a dimension of $(7 + 1)M \times 256$, linear layer 4 has a dimension of $256\times256$, and linear layer 5 has a dimension of $256\times512$. Linear layer 6 is a fully connected layer which outputs the state-value function.

Since the agents are heterogeneous in this application, they are trained in separate networks and no parameter sharing involves as all agents are non-exchangeable.

\subsection{Indirect MARL approach}\label{subsec:indirect_approach}
The transition probabilities between states needs to be provided to complete the MDP setting. The state transition probabilities, for DQJL application, are the collective result of the longitudinal and latitudinal dynamics, stochastic driving behavior and the coordination strategy.

The real road dynamics for partially connected settings are usually complicated and difficult to interpret. Hence, some fundamental models are incorporated to picture the road dynamics. The road dynamics can be categorized in the longitudinal and latitudinal direction taking account of the stochastic driving behavior as well as DQJL coordination.

\subsubsection{Longitudinal dynamics}
Along the direction of the traffic flow, it is natural to employ a car-following model to capture the longitudinal dynamics. In this study, we modify and adopt the discrete version of the intelligent driver model (IDM) \cite{gipps1981behavioural} so that vehicles' positions, velocities and accelerations can be computed across the discrete time horizon, i.e. MDP steps. To avoid denotation confusion, $u$ is used to represent acceleration.

For a non-EMV not in the yielding process, the discrete IDM model states for the next step
\begin{align*}
    v^{i}_{t+1} & = v^{i}_{t} + u^{i}_{t}\Delta t,\\
    x^{i}_{t+1} & = x^{i}_{t} + v^{i}_{t}\Delta t + \frac{u^{i}_{t}\Delta t}{2},
\end{align*}
in which the acceleration of the vehicle at step $t$ can be determined by the ego vehicle $i$ and its leading vehicle $j$ as
\begin{equation*}
    u^{i}_{t} = u_{0}[1 - (\frac{v^{i}_{t}}{v_{0}})^4 - (\frac{s^{*}(v^{i}_{t}, v^{j}_{t} - v^{i}_{t})}{x^{j}_{t} - x^{i}_{j} - l^{j}})^2],
\end{equation*}
where $s^{*}(v^{i}_{t}, v^{j}_{t} - v^{i}_{t})$ stands for the desired dynamic distance of two neighboring vehicles and can be approximated by
\begin{equation*}
    s^{*}(v^{i}_{t}, v^{j}_{t} - v^{i}_{t}) = d + v^{i}_{t}T + \frac{v^{i}_{t}(v^{j}_{t} - v^{i}_{t})}{2\sqrt{u_{0}b_{0}}}.
\end{equation*}

In the equations above, $u_{0}, b_{0}$ represents the IDM acceleration and deceleration baseline correspondingly, $T$ stands for the safety headway and $d$ represents the minimum safety gap between two neighboring vehicles. 

For a non-EMV in the yielding process, its longitudinal dynamics involves mainly deceleration. As illustrated in Fig. \ref{fig:braking}, there are two longitudinal uncertainties in a vehicle's braking motion. The first one represents drivers' perception-reaction time, $t_r$, which stands for a delay between human drivers' perception and execution. As suggested in \cite{McGehee2000} on drivers' perception abilities, the perception-reaction time follows the normal distribution of $t_r \sim \mathcal{N}(2.25s, (0.5s)^{2})$ is selected for this study. Decision time is also consider included in the perception-reaction time in this application as most decisions are made by the centralized controller. To realize the perception-reaction time, additional data structure is used to count the MDP step until the drivers start to decelerate.

The second type of uncertainty in the longitudinal direction is the deceleration of vehicles when they are yielding. Each vehicle has its unique baseline deceleration rate and the human driver has its own unique braking behavior. The deceleration of the vehicle is captured as a white noise in addition to the baseline deceleration rate $b^{*}$. Mathematically speaking, the deceleration, not counting for car-following model, for a vehicle after receiving an yielding instruction at time $T_{i}$ is:
\begin{align*}
    b^{i}_{t} = \begin{cases}
    0, & t \in [T_{i}, T_{i} + t_{r}).\\
    {b}^{*}_{i} + \epsilon_{decel}, & t \geq T_{i} + t_{r}.\\
\end{cases}
\end{align*}

Thus, the overall longitudinal dynamics for a non-EMV is described as a function of an vehicle's velocity, position, as well as its leading velocity and position:
\begin{align*}
    v^{i}_{t+1} = f(v^{i}_{t}, x^{i}_{t}, v^{ij}_{t}, x^{ij}_{t}) =\begin{cases}
    v^{i}_{t} + u^{i}_{t}\Delta t, & \text{if } \xi^{i}_{t} = 0,\\
    v^{i}_{t} - {b}^{i}_{t}\Delta t, & \text{ otherwise},\\
\end{cases}
\end{align*}
where $ij$ denotes the leading vehicle of vehicle $i$.

Notice that the EMV strictly follows the longitudinal dynamics. Its front bump position, velocity, and acceleration are determined by the distance between itself and its leading vehicle as well as the difference between their speeds. Since the maximum allowable speed of the EMV is much larger than that of a non-EMV, it is expected that the EMV travels faster when it doesn't have any leading vehicles, i.e. DQJL established.

\subsubsection{Latitudinal dynamics}
Inspired by the benchmark lane-changing model \cite{MOBIL_lane_changing}, the dynamic lane position of vehicles in a DQJL process are viewed to be the result of stochastic driving behaviors and coordination strategies. If an EMV is approaching on the passing lane of this road segment, the centralized controller aims to form a vehicle platoon on the neighboring lane. If a non-EMV on the passing lane starts to yield, it no longer follows the car-following model and begins to decelerate until it pulls over onto the neighboring lane, at which step this non-EMV stops yielding.

Meanwhile, a vehicle on the neighboring lane will also decelerate temporarily. The yielding process for this decelerating non-EMV is considered completed when it has a new leading vehicle in the platoon. All vehicles will not necessarily brake until stop. Thus, the yielding status $\xi^{i}$ can be tracked as
\begin{equation*}
\xi^{i}_{t+1} = g(\xi^{i}_{t}, a^{i}_{t}) = \begin{cases}
    1, \text{if } a^{i}_{t} = 1 \text{ or }  \xi^{i}_{t} = 1\\
    0, \text{otherwise}.
    \end{cases}
\end{equation*}

After the perception-reaction time, the yielding non-EMV attempts to pull over onto the neighboring lane, the probability of successfully lane changing, without considering for collisions, is captured as a geometric distribution parameter $p$. If the average time for successfully lane changing is $t_{lc}$ and temporal step length is $\Delta$, it is easy to derive that
\begin{equation*}
    p = \frac{1}{t_{lc}/\Delta t} \Rightarrow p = \frac{\Delta t}{t_{lc}},
\end{equation*}
and the lane position $y$ of a pulling-over vehicle for the next step can be determined as
\begin{equation*}
y^{i}_{t+1} = h(y^{i}_{t}, \xi^{i}_{t})
= \begin{cases}
0, & t \in [T_{i}, T_{i} + t_{r}),\\
Y \sim G(\frac{\Delta t}{t_{lc}}), & t \geq  t_{r}.\\
\end{cases}
\end{equation*}
\subsubsection{State transition for indirect MARL}
Adopting both longitudinal and latitudinal dynamics with stochastic driving behavior, it is easy to mathematically describe the next state $\mathbf{s}_{t+1} \sim \mathbf{s}_{t}$ with some joint action $\mathbf{a}$ to be
\begin{equation}\label{eq:kinematic_model}
    \mathbf{s}_{t+1}
    =\begin{bmatrix}
    x^{1}_{t} + v^{1}_{t}\Delta t& h(y^{1}_{t}, \xi^{1}_{t})& f(v^{1}_{t}, x^{1}_{t}, v^{1j}_{t}, x^{1j}_{t})& g(\xi^{i}_{t}, a^{i}_{t}) & l^{1} & b^{*}_{1} & \phi^{1}\\
    x^{2}_{t} + v^{2}_{t}\Delta t& h(y^{2}_{t}, \xi^{2}_{t})& f(v^{2}_{t}, x^{2}_{t}, v^{2j}_{t}, x^{2j}_{t})& g(\xi^{i}_{t}, a^{i}_{t}) & l^{2} & b^{*}_{2} & \phi^{2}\\
    \vdots & & & & & & \vdots\\
    x^{n}_{t} + v^{n}_{t}\Delta t& h(y^{n}_{t}, \xi^{n}_{t})& f(v^{n}_{t}, x^{n}_{t}, v^{nj}_{t}, x^{nj}_{t})& g(\xi^{i}_{t}, a^{n}_{t}) & l^{n} & b^{*}_{n} & \phi^{n}\\
    \end{bmatrix},
\end{equation}
where $ij$ represents $i$th vehicle's leading vehicle.

\subsection{Direct MARL approach}\label{subsec:direct_approach}
However, the real traffic conditions are much more complicated than the longitudinal and latitudinal dynamics. Human drivers are obeying car-following models, lane changing models, and other explicit or implicit driving dynamics with uncertainties. It is inefficient to capture all potential dynamics and develop a kinematics models to process through the proposed MARL framework. Thus, the proposed methodology should not assume a differentiable model of the road environment dynamics. Data-driven reinforcement learning is able to detect and understand the comprehensive road environment dynamics and the noise distributions, which represent stochastic driving behavior from human drivers. 

Furthermore, the proposed framework is compatible with various types of communication structures among the vehicles and the infrastructure. The CTDE framework cannot only be adopted to realize cooperative connected vehicles applications such as establishing DQJLs for EMVs, but also can be utilized for other connected vehicles applications, cooperative or competitive, particularly in those with stochastic dynamics incurred by human drivers.

\section{Simulation Test Bed Experimental Design}\label{sec:experiment}
Since it is impossible to perform in-field vehicle coordination training due to safety and expense, we employ Simulation of Urban Mobility (SUMO) \cite{SUMO2018} as the test bed. In this section, we demonstrate the experimental design with generating simulation data in Subsec. \ref{subsec:data_synthesis}, designing and implementing the simulation in Subsec. \ref{subsec:embeding}. Lastly, the model parameters and training hyperparameters are shown in Subsec. \ref{subsec:model_parameters}.
\begin{figure}
    \centering
    \includegraphics[width=\textwidth]{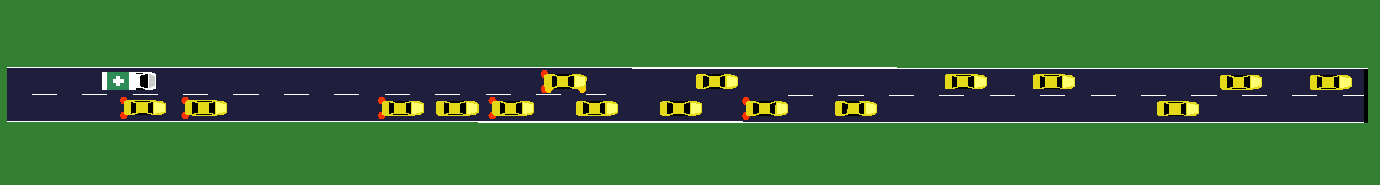}
    \caption{EMV is entering the road section from the left.}
    \label{fig:sumo_snapshot}
\end{figure}
\subsection{Data synthesis}\label{subsec:data_synthesis}
\begin{figure}[ht]
\begin{subfigure}{.5\textwidth}
  \centering
  \includegraphics[width=\linewidth]{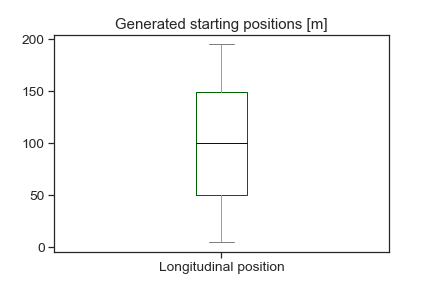}  
  \caption{Generated starting positions}
  \label{fig:staring_positions}
\end{subfigure}
\begin{subfigure}{.5\textwidth}
  \centering
  \includegraphics[width=\linewidth]{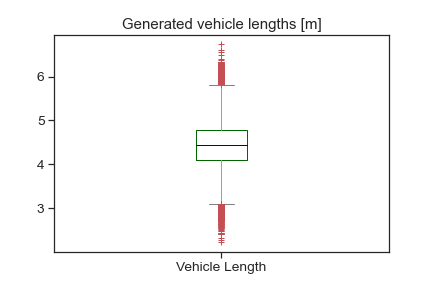}  
  \caption{Generated vehicle lengths}
  \label{fig:vehicle_lengths}
\end{subfigure}
\caption{The generated vehicle starting positions and average vehicle lengths.}
\label{fig:Data_summary}
\end{figure}

Providing realistic numerical values to simulate the realistic traffic conditions on a two-lane urban roadway is crucial for the reproduciblility of application in field. Therefore, a dataset containing randomly generate starting conditions, including different number of vehicles and different penetration ratio, is prepared to serve as the training set for both indirect and direct MARL approaches. To better compare the quantitative results under different penetration ratios, different number of non-EMVs are selected to further generate a grouped test set to reflect the difference under the chosen road environment configurations.\footnote{The dataset is available at \href{shorturl.at/PRT45}{shorturl.at/PRT45}.}
As shown in Table. \ref{tab:vehicle_feature}, we generate the vehicle features based on listed literature source. The statistics of some vehicle features are presented in Fig. \ref{fig:Data_summary}.
\begin{table}[!ht]
	\caption{Randomly generated vehicle features}\label{tab:vehicle_feature}
	\begin{center}
		\begin{tabular}{l l l l}
			Vehicle Feature & Value & Source\\\hline
			Vehicle Length $l$ & $\mathcal{N}(4.5m, (1m)^2)$ & \cite{Vehicle_length}\\
			Vehicle Baseline Deceleration $b^{*}$ & $ \mathcal{N}(-2m/s^2, (1m/s^2)^2)$ & \cite{BOKARE20174733, maurya2012study}\\
			Standard Deviation of Deceleration $\epsilon_{decel}$ & $0.5m/s^{2}$ & 
			\cite{BOKARE20174733, maurya2012study}\\
			\hline
		\end{tabular}
	\end{center}
\end{table}

The road segment studied is an arterial class III with LOS category C according to Arterial LOS standards \cite{roess2004traffic}. The nominal value for road environment configuration applied to both indirect and direct MARL training is listed in Table. \ref{tab:road_environment}. 
\begin{table}[!ht]
	\caption{System configuration parameters}\label{tab:road_environment}
	\begin{center}
		\begin{tabular}{l l l l}
			System Configuration Parameters & Value \\\hline
			Temporal Step Length $\Delta t$ & 0.5s\\
			Length of the Road segment $L$ & 200m\\
			Minimum Safety Gap $d$ & 0.5m\\
			Length of Reaction Distance $L_{HV}$ & 75m\\
			Non-EMV Starting Velocity $v_{0}^{1}$ & 4.5m/s\\
			EMV Starting Velocity $v_{0}^{0}$ & 8m/s\\
			EMV Maximum Allowable Velocity $v_{max}^{0}$ & 12m/s\\
			\hline
		\end{tabular}
	\end{center}
\end{table}

\subsection{RL-SUMO embedding}\label{subsec:embeding}
\begin{figure}
    \centering
    \includegraphics[width=\textwidth]{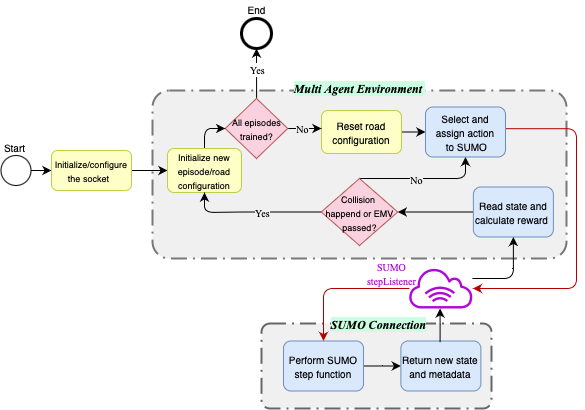}
    \caption{SUMO-based multi-agent reinforcement learning training pipeline for DQJL problem. The pipeline combines the multi-agent road environment with the SUMO test bed. The proposed pipeline processes and manages the observations, actions, and rewards as input or output to the SUMO platform. Coordinated by TraCI, the pipeline can initialize, load and reset experiment road configurations.}
    \label{fig:simulation_pipeline}
\end{figure}

For the indirect MARL approach, customized IDM parameters selected for the training process is listed in Table. \ref{tab:idm_parameters}. The trained coordination strategy is then validated on the test set on SUMO.
\begin{table}[!ht]
	\caption{Indirect MARL IDM model parameters}\label{tab:idm_parameters}
	\begin{center}
		\begin{tabular}{l l l l}
			Modified IDM model parameters & Value & Source\\\hline
			Baseline Acceleration $u_{0}$  & $3m/s^{2}$ & \cite{Treiber_2000}\\
			Baseline Deceleration $b_{0}$  & $-2m/s^{2}$ & \cite{Treiber_2000}\\
			Desired Traffic Speed $v^{*}$ & $10m/s$ &
			\cite{Treiber_2000, Kesting_2010}\\
			Desired Headway $T_{0}$ & $1.5s$ &
			\cite{Treiber_2000, Kesting_2010}\\
			\hline
		\end{tabular}
	\end{center}
\end{table}
For the direct MARL approach, the state transitions are completely determined by the SUMO dynamics. Instead of developing an accurate road dynamic model when yielding to approach EMVs, this approach does not assume an differentiable model and is able to handle all potential state transitions. Inspired by Flow \cite{SUMO2018:Flow_Deep_Reinforcement_Learning}, the embedding on SUMO is achieved via socket programming and the TraCI package in SUMO. The complete multi-agent reinforcement learning pipeline on SUMO is presented in Fig. \ref{fig:simulation_pipeline}.
\FloatBarrier

Instead of adopting the blue-light device, in SUMO sub-lane modulo, which provide a virtual passage between lanes \cite{bieker2018analysis, obrusniksimulating}, we initialize another vehicle starting upstream to serve as the EMV in the simulation. The EMV's initial velocity and maximum allowable velocity are set according to Table. \ref{tab:road_environment} for the fast passage through the segment. The other EMV vehicle features, such as vehicle length, are trivial as they do not influence the learning process.

In the direct MARL approach, built-in vehicle lane-changing models are overwritten so that non-EMVs only change lane when instructed, or forced if they are HVs. see Table. \ref{tab:LaneChanging}.
\begin{table}[!ht]
 \caption{Configuration parameters for SUMO vehicle lane-changing}\label{tab:LaneChanging}
 \begin{center}
  \begin{tabular}{l l l l}
   Lane Changing Parameters & Value \\\hline
   collision.mingap-factor & 0\\
   speed mode  &  0\\
   lane change mode & 0\\
   lsStrategic & 0 \\
   lsCooperative & 0.5\\
   \hline
  \end{tabular}
 \end{center}
\end{table}
\FloatBarrier

Similar to the noise perturbation adopted in the indirect MARL approach, identical distributions of noises reflecting stochastic driving behaviors are added to each vehicle during the training stage for the indirect MARL. Through TraCI, additional data structures are initialized to track and execute the randomness. For instance, if an drivers' perception reaction time is sampled as $2.5s$, a step counter is employed to make the driver move at $\frac{2.5s}{\Delta t} = 5$ steps delayed according to Table. \ref{tab:road_environment}. 

\subsection{Training hyperparameters}\label{subsec:model_parameters}
To maintain the consistency in training under indirect and direct MARL approaches, the training hyperparameters presented in Table. \ref{tab:RL_parameters} are selected for the MARL training stage.
\begin{table}[!ht]
	\caption{RL training hyperparameters}\label{tab:RL_parameters}
	\begin{center}
		\begin{tabular}{l l l l}
			Model Parameters and Hyper-parameters & Value \\\hline
			Priority Penalize Constant $P_{priority}$ & 0.5\\
		    Collision Penalize Constant	$P_{collision}$ & -1000\\
			Discount Factor $\gamma$ &  0.99\\
			Minibatch Size & 64\\
			Actor Learning Rate $\alpha$ & $10^{-4}$\\
			Critic Learning Rate $\beta$ & $10^{-3}$\\
			Replay Memory Size $D$ & 10000\\
			Initial Epsilon $\epsilon$ & 0.99\\
			Epsilon decay & $10^{-3}$\\
			Loss Optimizer & Adam\\
			\hline
		\end{tabular}
	\end{center}
\end{table}

\section{Validation Results and Discussion}\label{sec:validation}
In this section, we analyze the test bed validation results from the designed experiments.  Training performance for both indirect and direct MARL approaches with the proposed neural network structure are compared in Subsec. \ref{subsec:training_performance}. The validated EMV passing time resulted from both approaches are evaluated in Subsec. \ref{subsec:passing_time}, and their corresponding training time are assessed in Subsec. \ref{subsec:training_time}. The real-time coordination compatibility is also justified in Subsec. \ref{subsec:compatibility}.

\subsection{Training performance comparison}\label{subsec:training_performance}
\begin{figure}[ht]
\begin{subfigure}{.5\textwidth}
  \centering
  \includegraphics[width=\linewidth]{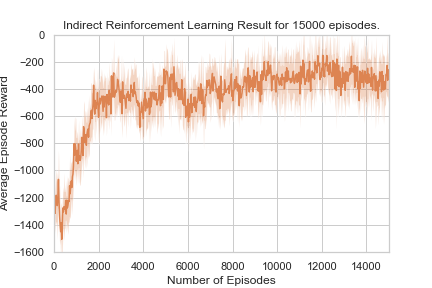}  
  \caption{Indirect MARL results after 15000 episodes.}
  \label{fig:Indirect_RL_result}
\end{subfigure}
\begin{subfigure}{.5\textwidth}
  \centering
  \includegraphics[width=\linewidth]{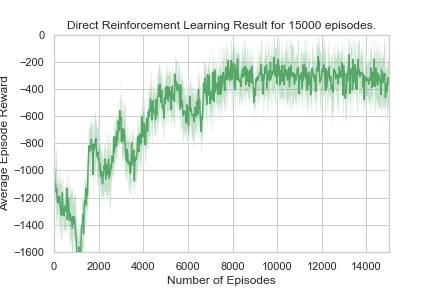}  
  \caption{Direct MARL results after 15000 episodes.}
  \label{fig:Direct_RL_result}
\end{subfigure}
\caption{The training performance for Fig. \ref{fig:Indirect_RL_result}-indirect MARL and Fig. \ref{fig:Direct_RL_result}-direct MARL.}
\label{fig:RL_learning_results}
\end{figure}

The learning performance with multiple independent runs for the indirect and direct reinforcement learning framework is presented in Fig. \ref{fig:RL_learning_results}. 
The dark lines highlights the mean value of these runs and the shaded area stands for one standard deviation.

According to the learning behaviors of the indirect and direct MARL approaches, the indirect MARL approach converges much faster w.r.t number of episodes. The indirect MARL approach achieves optimal average reward within approximately 3000 episodes, while the direct MARL approach converges at an approximately equivalent level within around 8000 episodes. The fast convergence of the indirect MARL approach takes advantage of the significant reduction in exploration time due to establishment of road environment dynamics beforehand. Both learning curves become stable onward, and the fluctuations representing the effects of randomness in road environment initialization and stochastic road dynamics.

\subsection{Emergency vehicle passing time validation}\label{subsec:passing_time}                                   
\begin{figure}
    \centering
    \includegraphics[width=\textwidth]{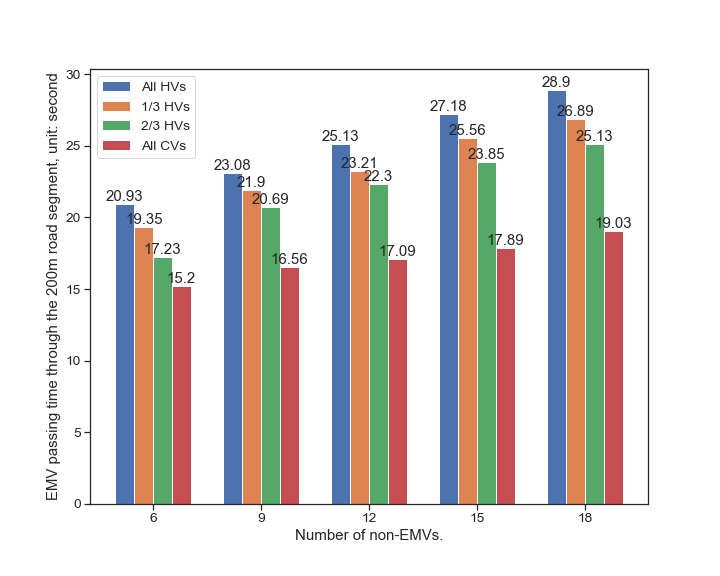}
    \caption{EMV passing time results from indirect reinforcement learning approach.}
    \label{fig:indirect_EMV_passing}
\end{figure}

To assess the DQJL coordination strategy from the proposed approaches, different groups of experiments are set up to compare the EMV passing time on a 200 meters road segment. The results under the indirect reinforcement learning framework are shown in Fig. \ref{fig:indirect_EMV_passing} grouped by different connectivity penetration rate. Within the same group of experiments, i.e. same number of non-EMVs, four scenarios with different penetration rates are separated but all other vehicles' features are kept consistent, including their starting positions.\footnote{An EMV passing demo is available at \href{shorturl.at/djsx4}{shorturl.at/djsx4} for qualitative analysis.}

According to Fig. \ref{fig:indirect_EMV_passing}, the EMV passing time exhibits monotonically decreasing pattern when the proportion of connected vehicles increases. The all HVs scenario is referred as the baseline scenario without any DQJL coordination. When the road is more crowded, it is evident from the data that distinctions of EMV passing time between all CVs and other scenarios broaden. The reason is the approaching EMV may be blocked by one more HV in the coordination process, significantly reducing its velocity meanwhile. As a result, the system performance of the DQJL coordination may be bottle-necked in the presence of HVs. 

Comparing the grouped results along the numbers of non-EMVs in the system, it appears that EMV passing time increments with the number of vehicles in the system. The increasing patterns differentiate, however, with different penetration rates. With all HVs in the traffic, the EMV passing time grows with an increasing manner when the road becomes more congested, whereas the EMV passing time with 100\% penetration rate grows marginally. It is then not difficult to imply that the difference will scale when the target road segment becomes increasing congested. Considering the results, with all connected vehicles, the proposed approach saves approximately 30\% of EMV passing time compared with the baseline scenario. The validation process is based on a Intel CPU i9 with 3.6 GHz with a NVIDIA GeForce 2080 Ti GPU, with an average decision time of 5.3 ms.
\begin{figure}
    \centering
    \includegraphics[width=\textwidth]{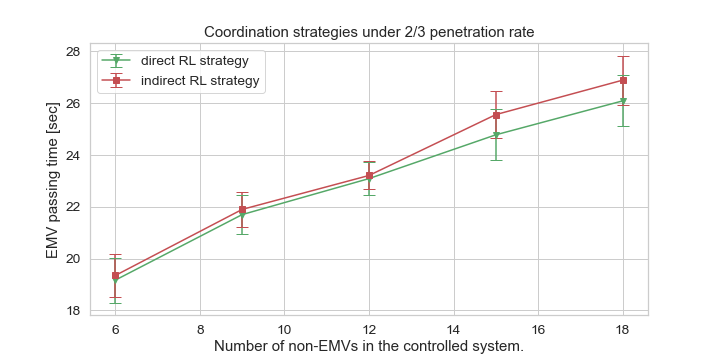}
    \caption{Comparison of EMV passing time between two approaches.}
    \label{fig:compared_EMV_passing_time}
\end{figure}

Correlating the EMV passing time with coordination strategies from the indirect MARL approach and direct MARL approach, the coordination strategies from these two approaches can be quantitatively assessed. As shown in Fig. \ref{fig:compared_EMV_passing_time} comparing the EMV passing time under coordination strategies from both approaches, it is straightforward to see that the coordination strategy from the direct MARL approach is slightly better than that from the indirect MARL approach as it results in shorter time of DQJL establishment, which can be explained by the fact that the direct MARL approach grasps the real testbed state transition better. Since the difference in EMV passing time between these two approaches is insignificant across densities, it is considered that both MARL approaches yield the optimal DQJL coordination strategies, especially when their learning curves converge at a comparable level.

\subsection{Training time comparison}\label{subsec:training_time}
The training time reflects the cost of the proposed reinforcement learning approaches. Although the proposed methodology is capable of handling various numbers of vehicles in the controlled system, a set of densities, representing a group of numbers of non-EMVs, is selected for the examination of the training time with both approaches. Instead of allowing both approaches converge at optimal reward level, training time for 5000 episodes are recorded.
\begin{figure}
    \centering
    \includegraphics[width=\textwidth]{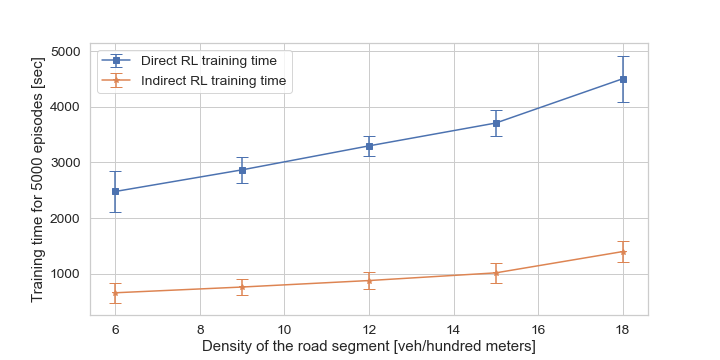}
    \caption{MARL training time for 5,000 episodes against different road densities.}
    \label{fig:training_time}
\end{figure}

Notice that the proposed methodology can handle various quantities of non-EMVs in the system and these densities are only chosen for experimenting the scalablity of the proposed methodologies. All training runs are conducted on a Intel CPU i9 with 3.6 GHz with a NVIDIA GeForce 2080 Ti GPU.

It is evident from the training time results shown in Fig.\ref{fig:training_time} that the direct reinforcement learning approach requires approximately 4 times of training time to finish 5000 episodes of training. The result is equivalent to say, the pipeline presented in Fig. \ref{fig:simulation_pipeline} of allocating, initializing, and exchanging information takes additional of 3 times of training time besides performing numerical computations. It is also noticeable that the training time is increasing with an accelerating pace, which implies the "curse of dimensionality" of the proposed algorithm. When the number of non-EMVs in the controlled system increases, the training time to reach optimal coordination tends to grow exponentially.
\subsection{Real-time compatibility validation}\label{subsec:compatibility}
The average decision time based on the validation configuration is 5.3ms. According to the 3GPP vehicle-to-everything communication standards \cite{3GPP}, the minimum requirement for an collision-free message sending interval is 10ms. Thus, the proposed framework is compatible for generating real-time coordination strategies with additional time for the support of functional expansions.
\section{Conclusion and Future Directions}\label{sec:conclusion}
In this study, we propose a multi-agent actor-critic based deep reinforcement algorithm to address the fast and safe DQJL establishment under the partially connected setting. By utilizing the centralized training with decentralized execution framework, the proposed methodology is scalable and extensible for multi-agent vehicular motion planning control problems in connected vehicle applications. In this study, two ways approaching the optimal coordination strategies are presented-indirect MARL approach based on an modified car-following model and stochastic driving behavior, and direct MARL approach based on the traffic simulation software. Both approaches are validated on the SUMO testbed, eventually, with the baseline scenario in which the traffic consists of all HVs and no coordination strategy is provided. The validation results suggest, with the real time coordination instructions informing the connected component in the traffic, the EMV passing time can be reduced by approximately 30\%, providing a promising direction for facilitating emergency service vehicles in intelligent urban road environment. Finally, we evaluate the training cost of both approaches and validate the real-time compatibility with the proposed connectivity schema. To summarize the contributions of the presented study, 
\begin{enumerate}
  \item we introduced the concept of dynamic queue-jump lane (DQJL) to reduce EMVs passing time on intelligent urban roadways;
  \item we developed a multi-agent actor-critic based deep reinforcement learning algorithm which addresses varying number of agent vehicles with discrete space for action representations for DQJL applications;
  \item we proposed a connected vehicle application framework which deals with partially connected traffic and stochastic driving behaviors;
  \item we presented two approaches, direct and indirect, when applying multi-agent reinforcement learning on connected vehicle applications and we validated the proposed coordination significantly improves emergency vehicle performances on connected urban roadway.
\end{enumerate}

The study can be extended into a few directions for the realization of the DQJL coordination in the field. First, regarding the privacy concern of a connected vehicle's policy through vehicular communications, policy inference networks can be additionally deployed onto each agent vehicle. The policy interference network allows each agent to learn other vehicles' policies without explicitly asking through communication channel. Second, the problem setting can be extended into signalized or unsignalized traffic intersections to recognize the multi-agent coordination pattern in a more complicated road environment. Realizing the intersection coordination and link-level coordination based on the presented work, we are able to present the network-level DQJL coordination, which will further improve the emergency vehicles' performance.
\section{Acknowledgment}
This research was supported by the C2SMART University Transportation Center at New York University, and Dwight David Eisenhower Transportation Fellowship Program (DDETFP) from the United States Department of Transportation. The authors thank Dr. Shuseng Wang for inspirations and advice on the methodology. The authors also thank Hengfeng Lu for preparing graphs.

\section{Declaration of Competing Interest}
The authors declare that they have no competing financial interests or personal relationships, to the best of their knowledge, that could have appeared to influence the work reported in this paper.

\bibliographystyle{elsarticle-num-names}
\bibliography{library.bib}

\begin{thebibliography}{52}
\expandafter\ifx\csname natexlab\endcsname\relax\def\natexlab#1{#1}\fi
\providecommand{\url}[1]{\texttt{#1}}
\providecommand{\href}[2]{#2}
\providecommand{\path}[1]{#1}
\providecommand{\DOIprefix}{doi:}
\providecommand{\ArXivprefix}{arXiv:}
\providecommand{\URLprefix}{URL: }
\providecommand{\Pubmedprefix}{pmid:}
\providecommand{\doi}[1]{\href{http://dx.doi.org/#1}{\path{#1}}}
\providecommand{\Pubmed}[1]{\href{pmid:#1}{\path{#1}}}
\providecommand{\bibinfo}[2]{#2}
\ifx\xfnm\relax \def\xfnm[#1]{\unskip,\space#1}\fi
\bibitem[{Government(2020{\natexlab{a}})}]{NY2019}
\bibinfo{author}{N.~Y.~C. Government}, \bibinfo{title}{New York End-To-End
  Response Times}, \bibinfo{year}{2019 (accessed November 22,
  2020)}{\natexlab{a}}. \URLprefix
  \url{https://www1.nyc.gov/site/911reporting/reports/end-to-end-repsonse-time.page}.
\bibitem[{Government(2020{\natexlab{b}})}]{Emergency2014}
\bibinfo{author}{N.~Y.~C. Government}, \bibinfo{title}{Emergency Response
  Incidents}, \bibinfo{year}{2014 (accessed February 28, 2020)}{\natexlab{b}}.
  \URLprefix
  \url{https://data.cityofnewyork.us/Public-Safety/Emergency-Response-Incidents/pasr-j7fb}.
\bibitem[{AHA(2020)}]{Heart2013}
\bibinfo{author}{AHA}, \bibinfo{title}{Heart Disease and Stroke Statistics},
  \bibinfo{year}{2013 (accessed February 28, 2020)}. \URLprefix
  \url{https://cpr.heart.org/AHAECC/CPRAndECC/ResuscitationScience/UCM_477263_AHA-Cardiac-Arrest-20Statistics.jsp5BR=301,L,NC5D}.
\bibitem[{Zhou and Gan(2005)}]{Zhou2005Performance}
\bibinfo{author}{G.~Zhou}, \bibinfo{author}{A.~Gan},
\newblock \bibinfo{title}{Performance of transit signal priority with queue
  jumper lanes},
\newblock \bibinfo{journal}{Transportation Research Record}
  \bibinfo{volume}{1925} (\bibinfo{year}{2005}) \bibinfo{pages}{265--271}.
  \DOIprefix\doi{10.1177/0361198105192500127}.
\bibitem[{Cesme et~al.(2015)Cesme, Altun, and Lane}]{Cesme2015Queue}
\bibinfo{author}{B.~Cesme}, \bibinfo{author}{S.~Z. Altun},
  \bibinfo{author}{B.~Lane},
\newblock \bibinfo{title}{Queue jump lane, transit signal priority, and stop
  location evaluation of transit preferential treatments using
  microsimulation},
\newblock \bibinfo{journal}{Transportation Research Record}
  \bibinfo{volume}{2533} (\bibinfo{year}{2015}) \bibinfo{pages}{39--49}.
  \DOIprefix\doi{10.3141/2533-05}.
\bibitem[{Buchenscheit et~al.(2009)Buchenscheit, Schaub, Kargl, and
  Weber}]{Buchenscheit2009AVE}
\bibinfo{author}{A.~Buchenscheit}, \bibinfo{author}{F.~Schaub},
  \bibinfo{author}{F.~Kargl}, \bibinfo{author}{M.~Weber},
\newblock \bibinfo{title}{A vanet-based emergency vehicle warning system},
\newblock \bibinfo{journal}{2009 IEEE Vehicular Networking Conference (VNC)}
  (\bibinfo{year}{2009}) \bibinfo{pages}{1--8}.
\bibitem[{Yasmin et~al.(2012)Yasmin, Anowar, and Tay}]{Yasmin2012Effects}
\bibinfo{author}{S.~Yasmin}, \bibinfo{author}{S.~Anowar},
  \bibinfo{author}{R.~Tay},
\newblock \bibinfo{title}{Effects of drivers’ actions on severity of
  emergency vehicle collisions},
\newblock \bibinfo{journal}{Transportation Research Record}
  \bibinfo{volume}{2318} (\bibinfo{year}{2012}) \bibinfo{pages}{90--97}.
  \DOIprefix\doi{10.3141/2318-11}.
\bibitem[{Savolainen et~al.(2010)Savolainen, Datta, Ghosh, and
  Gates}]{Savolainen2010Effects}
\bibinfo{author}{P.~T. Savolainen}, \bibinfo{author}{T.~K. Datta},
  \bibinfo{author}{I.~Ghosh}, \bibinfo{author}{T.~J. Gates},
\newblock \bibinfo{title}{Effects of dynamically activated emergency vehicle
  warning sign on driver behavior at urban intersections},
\newblock \bibinfo{journal}{Transportation Research Record}
  \bibinfo{volume}{2149} (\bibinfo{year}{2010}) \bibinfo{pages}{77--83}.
  \DOIprefix\doi{10.3141/2149-09}.
\bibitem[{{Hannoun} et~al.(2019){Hannoun}, {Murray-Tuite}, {Heaslip}, and
  {Chantem}}]{Hannoun2019Facilitating}
\bibinfo{author}{G.~J. {Hannoun}}, \bibinfo{author}{P.~{Murray-Tuite}},
  \bibinfo{author}{K.~{Heaslip}}, \bibinfo{author}{T.~{Chantem}},
\newblock \bibinfo{title}{Facilitating emergency response vehicles’ movement
  through a road segment in a connected vehicle environment},
\newblock \bibinfo{journal}{IEEE Transactions on Intelligent Transportation
  Systems} \bibinfo{volume}{20} (\bibinfo{year}{2019})
  \bibinfo{pages}{3546--3557}. \DOIprefix\doi{10.1109/TITS.2018.2877758}.
\bibitem[{Aslani et~al.(2019)Aslani, Mesgari, Seipel, and
  Wiering}]{aslani2019developing}
\bibinfo{author}{M.~Aslani}, \bibinfo{author}{M.~S. Mesgari},
  \bibinfo{author}{S.~Seipel}, \bibinfo{author}{M.~Wiering},
\newblock \bibinfo{title}{Developing adaptive traffic signal control by
  actor--critic and direct exploration methods},
\newblock in: \bibinfo{booktitle}{Proceedings of the Institution of Civil
  Engineers-Transport}, volume \bibinfo{volume}{172},
  \bibinfo{organization}{Thomas Telford Ltd}, \bibinfo{year}{2019}, pp.
  \bibinfo{pages}{289--298}.
\bibitem[{{Chu} et~al.(2020){Chu}, {Wang}, {Codecà}, and {Li}}]{Chu2020}
\bibinfo{author}{T.~{Chu}}, \bibinfo{author}{J.~{Wang}},
  \bibinfo{author}{L.~{Codecà}}, \bibinfo{author}{Z.~{Li}},
\newblock \bibinfo{title}{Multi-agent deep reinforcement learning for
  large-scale traffic signal control},
\newblock \bibinfo{journal}{IEEE Transactions on Intelligent Transportation
  Systems} \bibinfo{volume}{21} (\bibinfo{year}{2020})
  \bibinfo{pages}{1086--1095}. \DOIprefix\doi{10.1109/TITS.2019.2901791}.
\bibitem[{Guo et~al.(2019)Guo, Li, and {(Jeff) Ban}}]{GUO2019313}
\bibinfo{author}{Q.~Guo}, \bibinfo{author}{L.~Li}, \bibinfo{author}{X.~{(Jeff)
  Ban}},
\newblock \bibinfo{title}{Urban traffic signal control with connected and
  automated vehicles: A survey},
\newblock \bibinfo{journal}{Transportation Research Part C: Emerging
  Technologies} \bibinfo{volume}{101} (\bibinfo{year}{2019})
  \bibinfo{pages}{313 -- 334}. \URLprefix
  \url{http://www.sciencedirect.com/science/article/pii/S0968090X18311641}.
  \DOIprefix\doi{https://doi.org/10.1016/j.trc.2019.01.026}.
\bibitem[{Wei et~al.(2018)Wei, Zou, Zhang, Zhang, and Wang}]{wei2018design}
\bibinfo{author}{S.~Wei}, \bibinfo{author}{Y.~Zou}, \bibinfo{author}{T.~Zhang},
  \bibinfo{author}{X.~Zhang}, \bibinfo{author}{W.~Wang},
\newblock \bibinfo{title}{Design and experimental validation of a cooperative
  adaptive cruise control system based on supervised reinforcement learning},
\newblock \bibinfo{journal}{Applied sciences} \bibinfo{volume}{8}
  (\bibinfo{year}{2018}) \bibinfo{pages}{1014}.
\bibitem[{Yu et~al.(2018)Yu, Shao, Wei, and Zhou}]{yu2018intelligent}
\bibinfo{author}{L.~Yu}, \bibinfo{author}{X.~Shao}, \bibinfo{author}{Y.~Wei},
  \bibinfo{author}{K.~Zhou},
\newblock \bibinfo{title}{Intelligent land-vehicle model transfer trajectory
  planning method based on deep reinforcement learning},
\newblock \bibinfo{journal}{Sensors} \bibinfo{volume}{18}
  (\bibinfo{year}{2018}) \bibinfo{pages}{2905}.
\bibitem[{Yang et~al.(2017)Yang, Ozbay, and Ban}]{Yang2017Developments}
\bibinfo{author}{C.~Y.~D. Yang}, \bibinfo{author}{K.~Ozbay},
  \bibinfo{author}{X.~J. Ban},
\newblock \bibinfo{title}{Developments in connected and automated vehicles},
\newblock \bibinfo{journal}{Journal of Intelligent Transportation Systems}
  \bibinfo{volume}{21} (\bibinfo{year}{2017}) \bibinfo{pages}{251--254}.
  \URLprefix \url{https://doi.org/10.1080/15472450.2017.1337974}.
  \DOIprefix\doi{10.1080/15472450.2017.1337974}.
  \href{http://arxiv.org/abs/https://doi.org/10.1080/15472450.2017.1337974}{{\tt
  arXiv:https://doi.org/10.1080/15472450.2017.1337974}}.
\bibitem[{{Şahin} et~al.(2018){Şahin}, {Khalili}, {Boban}, and
  {Wolisz}}]{Sahin2018Reinforcement}
\bibinfo{author}{T.~{Şahin}}, \bibinfo{author}{R.~{Khalili}},
  \bibinfo{author}{M.~{Boban}}, \bibinfo{author}{A.~{Wolisz}},
\newblock \bibinfo{title}{Reinforcement learning scheduler for
  vehicle-to-vehicle communications outside coverage},
\newblock in: \bibinfo{booktitle}{2018 IEEE Vehicular Networking Conference
  (VNC)}, \bibinfo{year}{2018}, pp. \bibinfo{pages}{1--8}.
  \DOIprefix\doi{10.1109/VNC.2018.8628366}.
\bibitem[{Guan et~al.(2020)Guan, Ren, Li, Sun, Luo, and
  Li}]{guan2020centralized}
\bibinfo{author}{Y.~Guan}, \bibinfo{author}{Y.~Ren}, \bibinfo{author}{S.~E.
  Li}, \bibinfo{author}{Q.~Sun}, \bibinfo{author}{L.~Luo},
  \bibinfo{author}{K.~Li},
\newblock \bibinfo{title}{Centralized cooperation for connected and automated
  vehicles at intersections by proximal policy optimization},
\newblock \bibinfo{journal}{IEEE Transactions on Vehicular Technology}
  \bibinfo{volume}{69} (\bibinfo{year}{2020}) \bibinfo{pages}{12597--12608}.
\bibitem[{Su et~al.(2020)Su, Shi, Jin, and Chow}]{su2020v2i}
\bibinfo{author}{H.~Su}, \bibinfo{author}{K.~Shi}, \bibinfo{author}{L.~Jin},
  \bibinfo{author}{J.~Y.~J. Chow}, \bibinfo{title}{V2i connectivity-based
  dynamic queue-jump lane for emergency vehicles: A deep reinforcement learning
  approach}, \bibinfo{year}{2020}. \href{http://arxiv.org/abs/2008.00335}{{\tt
  arXiv:2008.00335}}.
\bibitem[{Tan(1993)}]{Tan93multi-agentreinforcement}
\bibinfo{author}{M.~Tan},
\newblock \bibinfo{title}{Multi-agent reinforcement learning: Independent vs.
  cooperative agents},
\newblock in: \bibinfo{booktitle}{In Proceedings of the Tenth International
  Conference on Machine Learning}, \bibinfo{publisher}{Morgan Kaufmann},
  \bibinfo{year}{1993}, pp. \bibinfo{pages}{330--337}.
\bibitem[{Puccetti et~al.(2019)Puccetti, Rathgeber, and
  Hohmann}]{puccetti2019actor}
\bibinfo{author}{L.~Puccetti}, \bibinfo{author}{C.~Rathgeber},
  \bibinfo{author}{S.~Hohmann},
\newblock \bibinfo{title}{Actor-critic reinforcement learning for linear
  longitudinal output control of a road vehicle},
\newblock in: \bibinfo{booktitle}{2019 IEEE Intelligent Transportation Systems
  Conference (ITSC)}, \bibinfo{organization}{IEEE}, \bibinfo{year}{2019}, pp.
  \bibinfo{pages}{2907--2913}.
\bibitem[{Huang et~al.(2017)Huang, Xu, He, Tan, and
  Sun}]{huang2017parameterized}
\bibinfo{author}{Z.~Huang}, \bibinfo{author}{X.~Xu}, \bibinfo{author}{H.~He},
  \bibinfo{author}{J.~Tan}, \bibinfo{author}{Z.~Sun},
\newblock \bibinfo{title}{Parameterized batch reinforcement learning for
  longitudinal control of autonomous land vehicles},
\newblock \bibinfo{journal}{IEEE Transactions on Systems, Man, and Cybernetics:
  Systems} \bibinfo{volume}{49} (\bibinfo{year}{2017})
  \bibinfo{pages}{730--741}.
\bibitem[{Lowe et~al.(2017)Lowe, Wu, Tamar, Harb, Abbeel, and
  Mordatch}]{lowe2017multiagent}
\bibinfo{author}{R.~Lowe}, \bibinfo{author}{Y.~Wu}, \bibinfo{author}{A.~Tamar},
  \bibinfo{author}{J.~Harb}, \bibinfo{author}{P.~Abbeel},
  \bibinfo{author}{I.~Mordatch}, \bibinfo{title}{Multi-agent actor-critic for
  mixed cooperative-competitive environments}, \bibinfo{year}{2017}.
  \href{http://arxiv.org/abs/1706.02275}{{\tt arXiv:1706.02275}}.
\bibitem[{Foerster et~al.(2017)Foerster, Farquhar, Afouras, Nardelli, and
  Whiteson}]{foerster2017counterfactual}
\bibinfo{author}{J.~Foerster}, \bibinfo{author}{G.~Farquhar},
  \bibinfo{author}{T.~Afouras}, \bibinfo{author}{N.~Nardelli},
  \bibinfo{author}{S.~Whiteson}, \bibinfo{title}{Counterfactual multi-agent
  policy gradients}, \bibinfo{year}{2017}.
  \href{http://arxiv.org/abs/1705.08926}{{\tt arXiv:1705.08926}}.
\bibitem[{Wu et~al.(2020)Wu, Jiang, and Zhang}]{wu2020cooperative}
\bibinfo{author}{T.~Wu}, \bibinfo{author}{M.~Jiang},
  \bibinfo{author}{L.~Zhang},
\newblock \bibinfo{title}{Cooperative multiagent deep deterministic policy
  gradient (comaddpg) for intelligent connected transportation with
  unsignalized intersection},
\newblock \bibinfo{journal}{Mathematical Problems in Engineering}
  \bibinfo{volume}{2020} (\bibinfo{year}{2020}).
\bibitem[{{Cao} et~al.(2020){Cao}, {Leng}, and {Zhang}}]{Cao2020Multi}
\bibinfo{author}{J.~{Cao}}, \bibinfo{author}{S.~{Leng}},
  \bibinfo{author}{K.~{Zhang}},
\newblock \bibinfo{title}{Multi-agent learning empowered collaborative decision
  for autonomous driving vehicles},
\newblock in: \bibinfo{booktitle}{2020 International Conference on UK-China
  Emerging Technologies (UCET)}, \bibinfo{year}{2020}, pp.
  \bibinfo{pages}{1--4}. \DOIprefix\doi{10.1109/UCET51115.2020.9205416}.
\bibitem[{Kwon and Kim(2019)}]{kwon2019multi}
\bibinfo{author}{D.~Kwon}, \bibinfo{author}{J.~Kim},
\newblock \bibinfo{title}{Multi-agent deep reinforcement learning for
  cooperative connected vehicles},
\newblock in: \bibinfo{booktitle}{2019 IEEE Global Communications Conference
  (GLOBECOM)}, \bibinfo{organization}{IEEE}, \bibinfo{year}{2019}, pp.
  \bibinfo{pages}{1--6}.
\bibitem[{Santa et~al.(2008)Santa, Gómez-Skarmeta, and
  Sánchez-Artigas}]{SANTA20082850}
\bibinfo{author}{J.~Santa}, \bibinfo{author}{A.~F. Gómez-Skarmeta},
  \bibinfo{author}{M.~Sánchez-Artigas},
\newblock \bibinfo{title}{Architecture and evaluation of a unified v2v and v2i
  communication system based on cellular networks},
\newblock \bibinfo{journal}{Computer Communications} \bibinfo{volume}{31}
  (\bibinfo{year}{2008}) \bibinfo{pages}{2850 -- 2861}.
  \DOIprefix\doi{https://doi.org/10.1016/j.comcom.2007.12.008},
  \bibinfo{note}{mobility Protocols for ITS/VANET}.
\bibitem[{Milanés et~al.(2010)Milanés, Godoy, Pérez, Vinagre, González,
  Onieva, and Alonso}]{MILANES201085}
\bibinfo{author}{V.~Milanés}, \bibinfo{author}{J.~Godoy},
  \bibinfo{author}{J.~Pérez}, \bibinfo{author}{B.~Vinagre},
  \bibinfo{author}{C.~González}, \bibinfo{author}{E.~Onieva},
  \bibinfo{author}{J.~Alonso},
\newblock \bibinfo{title}{V2i-based architecture for information exchange among
  vehicles},
\newblock \bibinfo{journal}{IFAC Proceedings Volumes} \bibinfo{volume}{43}
  (\bibinfo{year}{2010}) \bibinfo{pages}{85 -- 90}.
  \DOIprefix\doi{https://doi.org/10.3182/20100906-3-IT-2019.00017},
  \bibinfo{note}{7th IFAC Symposium on Intelligent Autonomous Vehicles}.
\bibitem[{Muhammad and Safdar(2018)}]{MUHAMMAD201850}
\bibinfo{author}{M.~Muhammad}, \bibinfo{author}{G.~A. Safdar},
\newblock \bibinfo{title}{Survey on existing authentication issues for
  cellular-assisted v2x communication},
\newblock \bibinfo{journal}{Vehicular Communications} \bibinfo{volume}{12}
  (\bibinfo{year}{2018}) \bibinfo{pages}{50 -- 65}.
  \DOIprefix\doi{https://doi.org/10.1016/j.vehcom.2018.01.008}.
\bibitem[{Scarzello et~al.(2001)Scarzello, Lenko, and
  Feaga}]{scarzello2001vehicle}
\bibinfo{author}{J.~F. Scarzello}, \bibinfo{author}{D.~S. Lenko},
  \bibinfo{author}{A.~C. Feaga}, \bibinfo{title}{Vehicle presence, speed and
  length detecting system and roadway installed detector therefor},
  \bibinfo{year}{2001}. \bibinfo{note}{US Patent 6,208,268}.
\bibitem[{Cheung et~al.(2005)Cheung, Coleri, Dundar, Ganesh, Tan, and
  Varaiya}]{cheung2005traffic}
\bibinfo{author}{S.~Y. Cheung}, \bibinfo{author}{S.~Coleri},
  \bibinfo{author}{B.~Dundar}, \bibinfo{author}{S.~Ganesh},
  \bibinfo{author}{C.-W. Tan}, \bibinfo{author}{P.~Varaiya},
\newblock \bibinfo{title}{Traffic measurement and vehicle classification with
  single magnetic sensor},
\newblock \bibinfo{journal}{Transportation Research Record}
  \bibinfo{volume}{1917} (\bibinfo{year}{2005}) \bibinfo{pages}{173--181}.
\bibitem[{Lazar et~al.(2019)Lazar, Bıyık, Sadigh, and
  Pedarsani}]{lazar2019learning}
\bibinfo{author}{D.~A. Lazar}, \bibinfo{author}{E.~Bıyık},
  \bibinfo{author}{D.~Sadigh}, \bibinfo{author}{R.~Pedarsani},
  \bibinfo{title}{Learning how to dynamically route autonomous vehicles on
  shared roads}, \bibinfo{year}{2019}.
  \href{http://arxiv.org/abs/1909.03664}{{\tt arXiv:1909.03664}}.
\bibitem[{{Bıyık} et~al.(2019){Bıyık}, {Lazar}, {Sadigh}, and
  {Pedarsani}}]{Biyik2019The}
\bibinfo{author}{E.~{Bıyık}}, \bibinfo{author}{D.~A. {Lazar}},
  \bibinfo{author}{D.~{Sadigh}}, \bibinfo{author}{R.~{Pedarsani}},
\newblock \bibinfo{title}{The green choice: Learning and influencing human
  decisions on shared roads},
\newblock in: \bibinfo{booktitle}{2019 IEEE 58th Conference on Decision and
  Control (CDC)}, \bibinfo{year}{2019}, pp. \bibinfo{pages}{347--354}.
  \DOIprefix\doi{10.1109/CDC40024.2019.9030169}.
\bibitem[{Bellman(1954)}]{bellman1954}
\bibinfo{author}{R.~Bellman},
\newblock \bibinfo{title}{The theory of dynamic programming},
\newblock \bibinfo{journal}{Bull. Amer. Math. Soc.} \bibinfo{volume}{60}
  (\bibinfo{year}{1954}) \bibinfo{pages}{503--515}.
\bibitem[{Konda and Tsitsiklis(2000)}]{konda2000actor}
\bibinfo{author}{V.~R. Konda}, \bibinfo{author}{J.~N. Tsitsiklis},
\newblock \bibinfo{title}{Actor-critic algorithms},
\newblock in: \bibinfo{booktitle}{Advances in neural information processing
  systems}, \bibinfo{year}{2000}, pp. \bibinfo{pages}{1008--1014}.
\bibitem[{Lillicrap et~al.(2019)Lillicrap, Hunt, Pritzel, Heess, Erez, Tassa,
  Silver, and Wierstra}]{lillicrap2019continuous}
\bibinfo{author}{T.~P. Lillicrap}, \bibinfo{author}{J.~J. Hunt},
  \bibinfo{author}{A.~Pritzel}, \bibinfo{author}{N.~Heess},
  \bibinfo{author}{T.~Erez}, \bibinfo{author}{Y.~Tassa},
  \bibinfo{author}{D.~Silver}, \bibinfo{author}{D.~Wierstra},
  \bibinfo{title}{Continuous control with deep reinforcement learning},
  \bibinfo{year}{2019}. \href{http://arxiv.org/abs/1509.02971}{{\tt
  arXiv:1509.02971}}.
\bibitem[{Haarnoja et~al.(2019)Haarnoja, Zhou, Hartikainen, Tucker, Ha, Tan,
  Kumar, Zhu, Gupta, Abbeel, and Levine}]{haarnoja2019soft}
\bibinfo{author}{T.~Haarnoja}, \bibinfo{author}{A.~Zhou},
  \bibinfo{author}{K.~Hartikainen}, \bibinfo{author}{G.~Tucker},
  \bibinfo{author}{S.~Ha}, \bibinfo{author}{J.~Tan},
  \bibinfo{author}{V.~Kumar}, \bibinfo{author}{H.~Zhu},
  \bibinfo{author}{A.~Gupta}, \bibinfo{author}{P.~Abbeel},
  \bibinfo{author}{S.~Levine}, \bibinfo{title}{Soft actor-critic algorithms and
  applications}, \bibinfo{year}{2019}.
  \href{http://arxiv.org/abs/1812.05905}{{\tt arXiv:1812.05905}}.
\bibitem[{Jang et~al.(2017)Jang, Gu, and Poole}]{gumbel_softmax}
\bibinfo{author}{E.~Jang}, \bibinfo{author}{S.~Gu}, \bibinfo{author}{B.~Poole},
  \bibinfo{title}{Categorical reparameterization with gumbel-softmax},
  \bibinfo{year}{2017}. \href{http://arxiv.org/abs/1611.01144}{{\tt
  arXiv:1611.01144}}.
\bibitem[{Gipps(1981)}]{gipps1981behavioural}
\bibinfo{author}{P.~G. Gipps},
\newblock \bibinfo{title}{A behavioural car-following model for computer
  simulation},
\newblock \bibinfo{journal}{Transportation Research Part B: Methodological}
  \bibinfo{volume}{15} (\bibinfo{year}{1981}) \bibinfo{pages}{105--111}.
\bibitem[{McGehee et~al.(2000)McGehee, Mazzae, and Baldwin}]{McGehee2000}
\bibinfo{author}{D.~V. McGehee}, \bibinfo{author}{E.~N. Mazzae},
  \bibinfo{author}{G.~S. Baldwin},
\newblock \bibinfo{title}{Driver reaction time in crash avoidance research:
  Validation of a driving simulator study on a test track},
\newblock \bibinfo{journal}{Proceedings of the Human Factors and Ergonomics
  Society Annual Meeting} \bibinfo{volume}{44} (\bibinfo{year}{2000})
  \bibinfo{pages}{3--320--3--323}. \DOIprefix\doi{10.1177/154193120004402026}.
  \href{http://arxiv.org/abs/https://doi.org/10.1177/154193120004402026}{{\tt
  arXiv:https://doi.org/10.1177/154193120004402026}}.
\bibitem[{Kesting et~al.(2007)Kesting, Treiber, and
  Helbing}]{MOBIL_lane_changing}
\bibinfo{author}{A.~Kesting}, \bibinfo{author}{M.~Treiber},
  \bibinfo{author}{D.~Helbing},
\newblock \bibinfo{title}{General lane-changing model mobil for car-following
  models},
\newblock \bibinfo{journal}{Transportation Research Record}
  \bibinfo{volume}{1999} (\bibinfo{year}{2007}) \bibinfo{pages}{86--94}.
  \DOIprefix\doi{10.3141/1999-10}.
  \href{http://arxiv.org/abs/https://doi.org/10.3141/1999-10}{{\tt
  arXiv:https://doi.org/10.3141/1999-10}}.
\bibitem[{Lopez et~al.(2018)Lopez, Behrisch, Bieker-Walz, Erdmann, Fltterd,
  Hilbrich, Lcken, Rummel, Wagner, and Wiener}]{SUMO2018}
\bibinfo{author}{P.~A. Lopez}, \bibinfo{author}{M.~Behrisch},
  \bibinfo{author}{L.~Bieker-Walz}, \bibinfo{author}{J.~Erdmann},
  \bibinfo{author}{Y.-P. Fltterd}, \bibinfo{author}{R.~Hilbrich},
  \bibinfo{author}{L.~Lcken}, \bibinfo{author}{J.~Rummel},
  \bibinfo{author}{P.~Wagner}, \bibinfo{author}{E.~Wiener},
\newblock \bibinfo{title}{Microscopic traffic simulation using sumo},
\newblock in: \bibinfo{booktitle}{The 21st IEEE International Conference on
  Intelligent Transportation Systems}, \bibinfo{publisher}{IEEE},
  \bibinfo{year}{2018}, p.~\bibinfo{pages}{1}.
\bibitem[{Coifman(2015)}]{Vehicle_length}
\bibinfo{author}{B.~Coifman},
\newblock \bibinfo{title}{Empirical flow-density and speed-spacing
  relationships: Evidence of vehicle length dependency},
\newblock \bibinfo{journal}{Transportation Research Part B: Methodological}
  \bibinfo{volume}{78} (\bibinfo{year}{2015}) \bibinfo{pages}{54 -- 65}.
  \DOIprefix\doi{https://doi.org/10.1016/j.trb.2015.04.006}.
\bibitem[{Bokare and Maurya(2017)}]{BOKARE20174733}
\bibinfo{author}{P.~Bokare}, \bibinfo{author}{A.~Maurya},
\newblock \bibinfo{title}{Acceleration-deceleration behaviour of various
  vehicle types},
\newblock \bibinfo{journal}{Transportation Research Procedia}
  \bibinfo{volume}{25} (\bibinfo{year}{2017}) \bibinfo{pages}{4733 -- 4749}.
  \DOIprefix\doi{https://doi.org/10.1016/j.trpro.2017.05.486},
  \bibinfo{note}{world Conference on Transport Research - WCTR 2016 Shanghai.
  10-15 July 2016}.
\bibitem[{Maurya and Bokare(2012)}]{maurya2012study}
\bibinfo{author}{A.~K. Maurya}, \bibinfo{author}{P.~S. Bokare},
\newblock \bibinfo{title}{Study of deceleration behaviour of different vehicle
  types.},
\newblock \bibinfo{journal}{International Journal for Traffic \& Transport
  Engineering} \bibinfo{volume}{2} (\bibinfo{year}{2012}).
\bibitem[{Roess et~al.(2004)Roess, Prassas, and McShane}]{roess2004traffic}
\bibinfo{author}{R.~P. Roess}, \bibinfo{author}{E.~S. Prassas},
  \bibinfo{author}{W.~R. McShane}, \bibinfo{title}{Traffic engineering},
  \bibinfo{publisher}{Pearson/Prentice Hall}, \bibinfo{year}{2004}.
\bibitem[{Treiber et~al.(2000)Treiber, Hennecke, and Helbing}]{Treiber_2000}
\bibinfo{author}{M.~Treiber}, \bibinfo{author}{A.~Hennecke},
  \bibinfo{author}{D.~Helbing},
\newblock \bibinfo{title}{Congested traffic states in empirical observations
  and microscopic simulations},
\newblock \bibinfo{journal}{Physical Review E} \bibinfo{volume}{62}
  (\bibinfo{year}{2000}) \bibinfo{pages}{1805–1824}.
  \DOIprefix\doi{10.1103/physreve.62.1805}.
\bibitem[{Kesting et~al.(2010)Kesting, Treiber, and Helbing}]{Kesting_2010}
\bibinfo{author}{A.~Kesting}, \bibinfo{author}{M.~Treiber},
  \bibinfo{author}{D.~Helbing},
\newblock \bibinfo{title}{Enhanced intelligent driver model to access the
  impact of driving strategies on traffic capacity},
\newblock \bibinfo{journal}{Philosophical Transactions of the Royal Society A:
  Mathematical, Physical and Engineering Sciences} \bibinfo{volume}{368}
  (\bibinfo{year}{2010}) \bibinfo{pages}{4585–4605}. \URLprefix
  \url{http://dx.doi.org/10.1098/rsta.2010.0084}.
  \DOIprefix\doi{10.1098/rsta.2010.0084}.
\bibitem[{Kheterpal et~al.(2018)Kheterpal, Parvate, Wu, Kreidieh, Vinitsky, and
  Bayen}]{SUMO2018:Flow_Deep_Reinforcement_Learning}
\bibinfo{author}{N.~Kheterpal}, \bibinfo{author}{K.~Parvate},
  \bibinfo{author}{C.~Wu}, \bibinfo{author}{A.~Kreidieh},
  \bibinfo{author}{E.~Vinitsky}, \bibinfo{author}{A.~Bayen},
\newblock \bibinfo{title}{Flow: Deep reinforcement learning for control in
  sumo},
\newblock in: \bibinfo{booktitle}{SUMO 2018- Simulating Autonomous and
  Intermodal Transport Systems}, volume~\bibinfo{volume}{2} of
  \textit{\bibinfo{series}{EPiC Series in Engineering}},
  \bibinfo{publisher}{EasyChair}, \bibinfo{year}{2018}, pp.
  \bibinfo{pages}{134--151}. \DOIprefix\doi{10.29007/dkzb}.
\bibitem[{Bieker-Walz et~al.(2018)Bieker-Walz, Behrisch, and
  Junghans}]{bieker2018analysis}
\bibinfo{author}{L.~Bieker-Walz}, \bibinfo{author}{M.~Behrisch},
  \bibinfo{author}{M.~Junghans},
\newblock \bibinfo{title}{Analysis of the traffic behavior of emergency
  vehicles in a microscopic traffic simulation},
\newblock \bibinfo{journal}{EPiC Series in Engineering} \bibinfo{volume}{2}
  (\bibinfo{year}{2018}) \bibinfo{pages}{1--13}.
\bibitem[{Obrusnik(2019)}]{obrusniksimulating}
\bibinfo{author}{V.~Obrusnik},
\newblock \bibinfo{title}{Simulating the impact of prioritization of emergency
  vehicles at traffic light controlled junctions on the other traffic},
\newblock \bibinfo{journal}{CTU in PRAGUE Master Thesis}
  (\bibinfo{year}{2019}).
\bibitem[{Wang et~al.(2017)Wang, Mao, and Gong}]{3GPP}
\bibinfo{author}{X.~Wang}, \bibinfo{author}{S.~Mao}, \bibinfo{author}{M.~Gong},
\newblock \bibinfo{title}{An overview of 3gpp cellular vehicle-to-everything
  standards},
\newblock \bibinfo{journal}{GetMobile: Mobile Computing and Communications}
  \bibinfo{volume}{21} (\bibinfo{year}{2017}) \bibinfo{pages}{19--25}.
  \DOIprefix\doi{10.1145/3161587.3161593}.

\end{thebibliography}





\end{document}